%% file: egocross.tex
\documentclass[letterpaper]{article} 
\usepackage{aaai2026}  
\usepackage{times}  
\usepackage{helvet}  
\usepackage{courier}  
\usepackage[hyphens]{url}  
\usepackage{graphicx} 
\usepackage{natbib}  
\usepackage{caption} 
\usepackage{algorithm}
\usepackage{algorithmic}
\usepackage{makecell}
\usepackage{array}
\usepackage{booktabs}
\usepackage{multirow}
\usepackage{cuted}

\usepackage{newfloat}
\usepackage{listings}
\DeclareCaptionStyle{ruled}{labelfont=normalfont,labelsep=colon,strut=off} 
\floatstyle{ruled}
\newfloat{listing}{tb}{lst}{}
\floatname{listing}{Listing}
\usepackage{subcaption}
\usepackage{amsmath}
\usepackage{amssymb}
\usepackage{pifont}
\usepackage{tabularx}
\usepackage{siunitx}

\usepackage[table]{xcolor}
\definecolor{summaryrow}{gray}{0.9}

\title{EgoCross: Benchmarking Multimodal Large Language Models for \\ Cross-Domain Egocentric Video Question Answering}

\author{
    Yanjun Li\textsuperscript{\rm 1}\equalcontrib,
    Yuqian Fu\textsuperscript{\rm 2}\equalcontrib,
    Tianwen Qian\textsuperscript{\rm 1}\protect\thanks{Corresponding authors.},
    Qi'ao Xu\textsuperscript{\rm 1},
    Silong Dai\textsuperscript{\rm 1},\\
    Danda Pani Paudel\textsuperscript{\rm 2},
    Luc Van Gool\textsuperscript{\rm 2},
    Xiaoling Wang\textsuperscript{\rm 1}\protect\footnotemark[2]
}
\affiliations{
    \textsuperscript{\rm 1}School of Computer Science and Technology, East China Normal University\\
    \textsuperscript{\rm 2}INSAIT, Sofia University ``St.~Kliment Ohridski''\\
}

\begin{document}

\maketitle

\begin{figure*}[!t] 
    \centering    \includegraphics[width=0.95\textwidth]{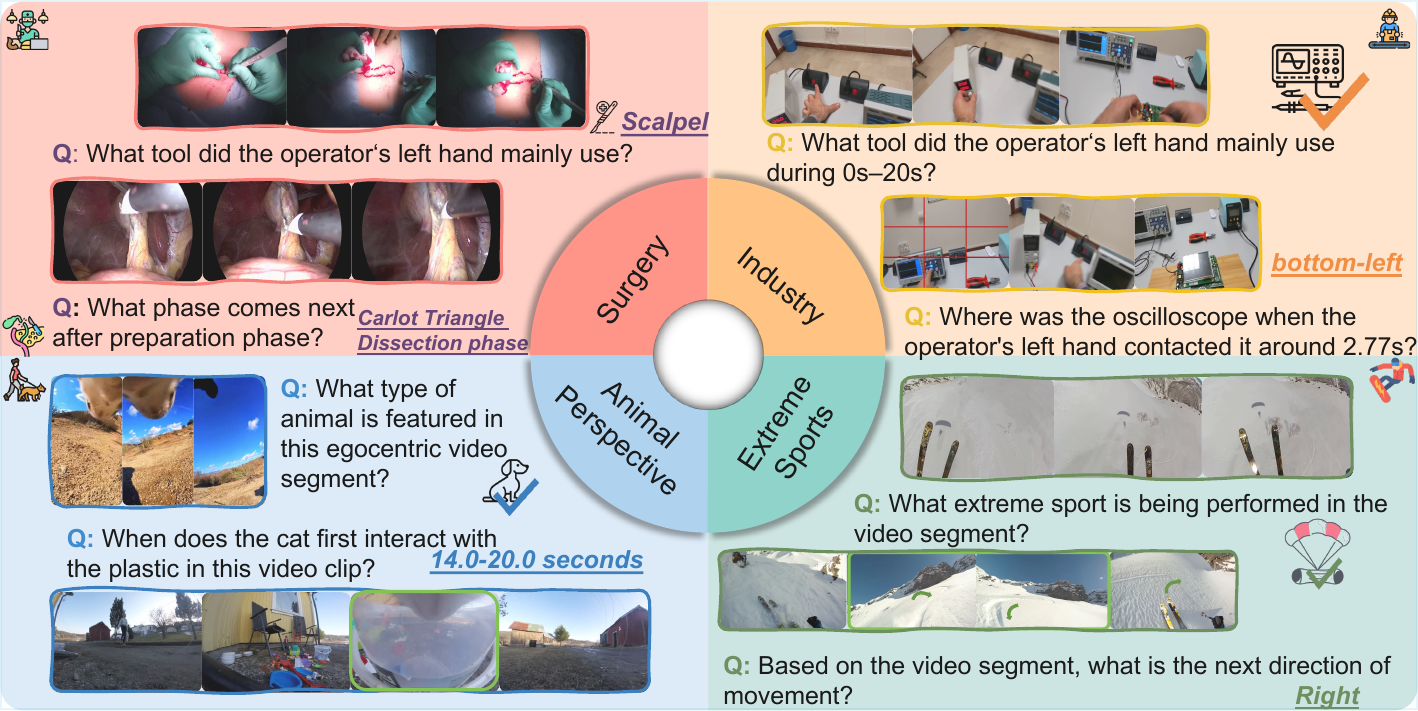}
   \caption{
   \textbf{Examples of Our EgoCross Benchmark.} We go beyond everyday egocentric scenarios, covering four diverse, cross-domain, application-oriented areas: Surgery, Industry, Extreme Sports, and Animal Perspective. As shown in the examples, both the visual appearances and the semantic content differ significantly from existing EgocentricQA datasets.
   }
    \label{fig:teaser}
\end{figure*}

\begin{abstract}

Recent advances in Multimodal Large Language Models (MLLMs) have significantly pushed the frontier of egocentric video question answering (EgocentricQA).
However, existing benchmarks and studies are mainly limited to common daily activities such as cooking and cleaning.
In contrast, real-world deployment inevitably encounters domain shifts, where target domains differ substantially in both visual style and semantic content.
To bridge this gap, we introduce \textbf{EgoCross}, a comprehensive benchmark designed to evaluate the cross-domain generalization of MLLMs in EgocentricQA.
EgoCross covers four diverse and challenging domains, including surgery, industry, extreme sports, and animal perspective, representing realistic and high-impact application scenarios.
It comprises approximately 1,000 QA pairs across 798 video clips, spanning four key QA tasks: prediction, recognition, localization, and counting.
Each QA pair provides both OpenQA and CloseQA formats to support fine-grained evaluation.
Extensive experiments show that most existing MLLMs, whether general-purpose or egocentric-specialized, struggle to generalize to domains beyond daily life, highlighting the limitations of current models.
Furthermore, we conduct several pilot studies, e.g., fine-tuning and reinforcement learning, to explore potential improvements.
We hope EgoCross and our accompanying analysis will serve as a foundation for advancing domain-adaptive, robust egocentric video understanding.

\end{abstract}

\begin{links}
\link{Code}{https://github.com/MyUniverse0726/EgoCross}
\link{Challenge}{https://egocross-benchmark.github.io/}
\end{links}

\section{Introduction}
Egocentric videos, which capture how humans perceive and interact with the physical world from a first-person perspective, offer a rich and unique source of data for modeling human behaviors. 
Understanding egocentric vision is therefore highly valuable for applications such as embodied AI, wearable assistants, and human-to-robot learning.
Among various egocentric tasks, video question answering (VQA)~\cite{zhong2022video} has emerged as a particularly challenging yet impactful problem.
Early efforts like EgoVQA~\cite{fan2019egovqa}, EgoTaskQA~\cite{jia2022egotaskqa}, and EgoSchema~\cite{mangalam2023egoschema} laid the groundwork for EgocentricQA by introducing dedicated benchmarks. The rapid progress of Multimodal Large Language Models (MLLMs) has further significantly advanced this field in both benchmark construction and model development. On the benchmark side, EgoThink~\cite{cheng2024egothink}, EgoTempo~\cite{plizzari2025omnia}, and EgoTextVQA~\cite{zhou2025egotextvqa} have been proposed, targeting different aspects of the QA task. On the modeling side, a number of MLLMs specifically designed or adapted for egocentric video understanding have also emerged. Notable examples include EgoVLPv2~\cite{pramanick2023egovlpv2} and EgoGPT~\cite{yang2025egolife}, which extend general-purpose MLLMs for EgocentricQA by training on specialized egocentric data.

Despite recent progress, most existing works remain focused on common daily-life activities, such as cooking, eating, and gardening. However, real-world applications inevitably extend beyond such scenarios. For example, in a surgical setting, a model must not only recognize a generic ``cutting tool" but also precisely differentiate between instruments like a grasper, a cautery hook, and bipolar forceps. In such cases, both the visual appearance and the semantic context deviate significantly from those found in everyday activities. This naturally raises a fundamental question: \textit{Can existing MLLMs generalize effectively to these uncommon and domain-specific scenarios?}

To answer this question, we introduce \textbf{EgoCross}, a comprehensive benchmark designed to evaluate the cross-domain generalization capabilities of MLLMs in EgocentricQA.
EgoCross is built upon three core design principles: \ding{172} emphasis on cross-domain properties, \ding{173} relevance to practical applications, 
and \ding{174} fine-grained, multi-dimensional model assessment.
Following these principles, we carefully curated video sources and developed corresponding QA pairs to reflect real-world, high-impact use cases.
Specifically, we selected surgery, industry, extreme sports, and animal perspective, as the four basic domains of our benchmark.
These domains exhibit substantial visual and semantic deviations from typical daily-life scenarios, thus posing unique challenges for model generalization.
Based on these video sources, we designed a structured data curation pipeline to construct QA pairs across four fundamental QA task types: \textit{identification}, \textit{localization}, \textit{prediction}, and \textit{counting}, further spanning a total of 15 specific subtasks.
To support both discriminative and generative evaluation protocols, each QA instance is annotated in both CloseQA (multiple-choice) and OpenQA (free-form answer) formats.
In total, EgoCross consists of approximately 1,000 QA pairs across 798 egocentric video clips, forming a carefully constructed dataset that enables systematic evaluation of cross-domain generalization in EgocentricQA.
A visual overview and representative examples are provided in Fig.~\ref{fig:teaser}.

Experiments demonstrate that most general-purpose and egocentric-specific MLLMs struggle on EgoCross, with CloseQA accuracy below 55\% (random chance: 25\%) and OpenQA below 35\%, revealing their limitations in cross-domain settings.
A notable performance drop (1.6$\times$$\downarrow$) on the same question types from EgoSchema to our EgoCross further confirms the challenge.
We also explored prompt learning, fine-tuning, and reinforcement learning to assess potential improvements, offering insights for future research.

Our main contributions are summarized as follows:
\begin{itemize}
\item 
We are the first to define and motivate the task of cross-domain EgocentricQA, an underexplored yet crucial area for real-world application.
\item 
We release EgoCross, the first cross-domain benchmark for EgocentricQA, covering four distinct domains (surgery, industry, extreme sports, and animal perspective) with $\sim$1k high-quality QA pairs.
\item 
We conduct a comprehensive evaluation across 8 state-of-the-art MLLMs, quantitatively revealing their limitations beyond daily-life domains and highlighting the need for more domain-robust models.
\item 
We provide forward-looking pilot studies, offering actionable insights and shedding light on future directions for building more generalizable and robust MLLMs.
\end{itemize}

\section{Related Work}
\subsection{Egocentric Video Understanding}
Egocentric video understanding, modeling human perception from a first-person view, has gained growing attention.Beyond basic perception~\cite{wang2021interactive, wang2023scene}, EgocentricQA has emerged as a key task.
Initial benchmarks like EgoVQA~\cite{fan2019egovqa}, EgoTaskQA~\cite{jia2022egotaskqa}, and EgoSchema~\cite{mangalam2023egoschema} have been joined by new datasets focusing on complex reasoning~\cite{cheng2024egothink}, temporal understanding~\cite{plizzari2025omnia}, and scene text~\cite{zhou2025egotextvqa}. The increasing data has also spurred the development of specialized models for egocentric video understanding, typically adapted from MLLMs. However, most existing work remains confined to daily-life scenarios, with limited attention paid to domain shifts. Our work fills this gap by introducing the first cross-domain testbed for EgocentricQA, emphasizing real-world, out-of-distribution targets.

\subsection{MLLMs for Video Understanding}
Recent MLLMs have shown remarkable capabilities in video understanding. General MLLMs such as GPT-4.1~\cite{achiam2023gpt}, Gemini 2.5 Pro~\cite{comanici2025gemini}, Qwen2.5-VL~\cite{bai2025qwen2}, and InternVL~\cite{zhu2025internvl3} achieve strong performance across a range of video tasks through extensive multimodal pretraining. In parallel, specialized models like Video-LLaMA3~\cite{zhang2025videollama} further improve temporal reasoning via dedicated architectural designs. Several MLLMs have also been tailored specifically for egocentric video understanding, including EgoVLPv2~\cite{pramanick2023egovlpv2} and EgoGPT~\cite{yang2025egolife}. While these models perform well on third-person videos and egocentric videos from common daily scenarios, their ability to generalize to unfamiliar, domain-specific scenarios remains largely unexamined. In this work, we systematically assess how well the current state-of-the-art MLLMs generalize to cross-domain egocentric targets, revealing their limitations and offering in-depth analysis to facilitate future research in this direction.

\subsection{Cross-Domain Generalization}

Cross-domain generalization is a broad and long-standing challenge in computer vision. Prior work has investigated it across various tasks, including image classification~\cite{zhu2019aligning, fu2023styleadv, zhao2020learning}, object detection~\cite{zheng2020cross, li2025domain, zhang2022divide}, and action recognition~\cite{pan2020adversarial, xu2022aligning, chen2021m3net}, often leveraging domain transfer, data augmentation, efficient fine-tuning, and meta-learning techniques~\cite{chen2021blockmix}. However, these efforts have primarily focused on third-person viewpoints and low-level perception tasks.
In egocentric video understanding, domain shifts are particularly pronounced due to drastic variations in scenes, task semantics, and camera motion. While some works explore cross-domain few-shot recognition in egocentric videos~\cite{hatano2024multimodal} or leverage large models for few-shot knowledge transfer~\cite{ge2023connecting}, these remain limited to low-level perception tasks or more general settings. In contrast, EgoCross is the first benchmark specifically designed to evaluate cross-domain generalization in EgocentricQA, tackling both domain gap and high-level reasoning challenges.


\section{EgoCross Dataset}

In this section, we provide a comprehensive introduction to the EgoCross benchmark.
We begin by discussing the selection of domains, video sources, and the taxonomy of question-answering tasks, followed by an explanation of the data curation pipeline, and conclude with it dataset statistics.

\subsection{Source Selection and Task Taxonomy}

\paragraph{Design Principles.}

We established key principles for domain and dataset selection, as well as question-answering task taxonomy: \textit{\ding{172} Emphasis on Cross-Domain Properties.} We need to select domains with distinct knowledge structures, terminologies, and interactions that differ significantly from everyday scenarios, ensuring the models are challenged by unfamiliar concepts. \textit{\ding{173} Impact on Practical Applications.} Datasets closely related to real-world applications, e.g., healthcare and industrial operations, are encouraged, as they are expected to foster progress toward practical applications of EgocentricQA.
\textit{\ding{174} Fine-grained Multi-dimensional Model Assessment.} Tasks should span a broad range, covering diverse examination types, such as complex reasoning and spatiotemporal dependencies, and also with comprehensive evaluation metrics.

\paragraph{Domain and Data Source Selection.}
Based on the above criteria, we select four professional domains that present distinct challenges and high real-world relevance: \textit{surgery}, \textit{industry}, \textit{extreme sports}, and \textit{animal perspective}.
For each, we curated one or two high-quality, open-source datasets with expert-provided meta annotations, each presenting unique perceptual, cognitive, and reasoning demands. The selected domains and the corresponding datasets are as follows:

\begin{itemize}
    \item \textbf{Surgery.}
    The surgical domain represents a highly structured, knowledge-intensive scenario where precision, sequential understanding, and risk-awareness are paramount.
    To enrich visual diversity, we include two datasets: \textit{EgoSurgery}~\cite{fujii2024egosurgeryphase}, which records the videos of open-heart surgeries from the surgeon's perspective, with fine-grained annotations of hand-tool interactions and surgical phases; and \textit{CholecTrack20}~\cite{nwoye2023cholectrack20}, which offers laparoscopic videos of cholecystectomy procedures from a tool-centered perspective. 
    This dataset offers a rare yet insightful egocentric perspective captured from a tool rather than a typical human operator.
    \item \textbf{Industry.}
    Complex workflows in industrial scenarios demand not only perception of fine object manipulations but also reasoning over procedural sequences and tool-usage logic.
    We choose \textit{ENIGMA-51}~\cite{ragusa2024enigma}, a dataset containing real circuit board repair tasks. 
    \item \textbf{Extreme Sports.}
    Extreme sports pose unique challenges, such as rare environments, rapid camera motion, and blur, which could well test models' spatiotemporal perception and high-speed situational reasoning.
    We include the \textit{ExtremeSportFPV}~\cite{singh2017trajectory}, which features first-person videos of various extreme sports, including mountain biking, skiing, and skydiving.
    \item \textbf{Animal Perspective.}
    To challenge anthropocentric bias in existing models, we introduce the animal perspective, introducing new motion patterns, camera angles, and semantic focus to the models.
    \textit{EgoPet}~\cite{bar2024egopet}, a dataset featuring egocentric views from animals such as dogs, cats, eagles, and turtles, is thus included.
\end{itemize}

The selected domains and video sources align well with our principles \ding{172} and \ding{173}.

\begin{figure*}[t]
    \centering
    \includegraphics[width=\textwidth]{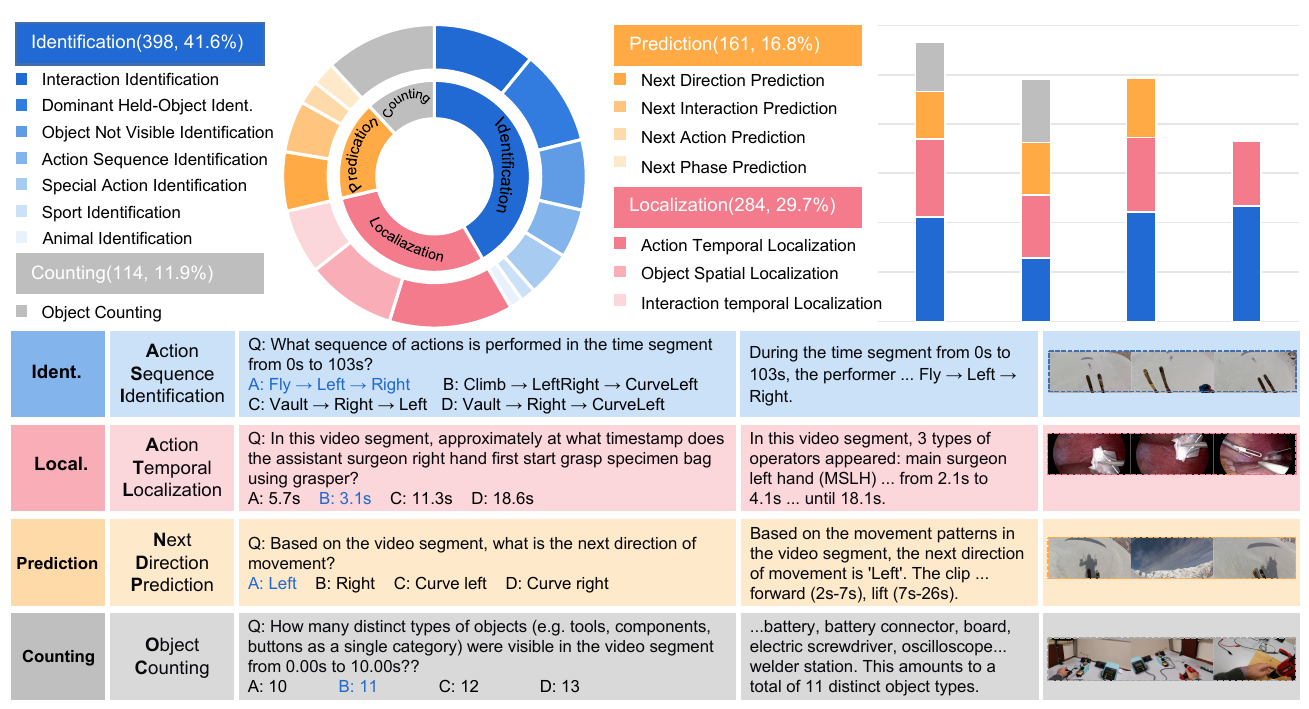}
    \caption{
        An overview of the EgoCross task taxonomy and statistics. (Top-left) The overall distribution of the four main task categories: Identification, Localization, Prediction, and Counting.
        (Top-right) The number of questions across the four primary domains. (Bottom) A selection of representative QA examples for each major capability is presented.
        For a more comprehensive list of examples, please see the Appendix.}
    \label{fig:task_taxonomy}
\end{figure*}

\paragraph{QA Task Taxonomy.}
Following Principle \ding{174}, we aim to construct diverse QA pairs to comprehensively assess model capabilities.
As illustrated in Fig.~\ref{fig:task_taxonomy}, our evaluation framework is built around four core task categories: \textit{Identification}, \textit{Localization}, \textit{Prediction}, and \textit{Counting}.
Tailored to address the unique challenges of each domain, we further decompose these four broad categories into 15 specific sub-tasks, collectively forming a comprehensive evaluation framework.
In the following, we provide an overview of the 4 core tasks.
Detailed sub-tasks and representative examples can be found in Fig.~\ref{fig:task_taxonomy} and Appendix.

\begin{itemize}
    \item \textbf{Identification.}
    Identification tasks evaluate a model's ability to recognize objects, actions, and events within a video. These tasks require domain-specific knowledge and adaptation to subtle differences in object properties or actions across contexts.
    For example, in surgical scenarios, instruments like forceps and scissors share similar shapes and colors, placing high demands on the models.
    \item \textbf{Localization.}
    Localization tasks assess a model's ability to identify the precise spatial or temporal location of objects, actions, or interactions.
    These tasks require the model to understand spatiotemporal relationships and adjust to the variations in object positioning and motion patterns across different environments.
    In industrial assembly, for example, locating tools or components in a cluttered workspace is particularly challenging due to partial occlusions and rapid movements of small objects.
    \item \textbf{Prediction.}
    Prediction tasks test a model's ability to forecast future actions or outcomes based on the current content.
    Thus, models are expected to grasp the underlying procedural or causal relationships in unseen domains.
    For example, in surgery, predicting the next phase of the procedure, such as transitioning from suturing to wound closure, relies on models' understanding of established surgical patterns, which can vary significantly across different domains like industrial assembly or extreme sports.
    \item \textbf{Counting.}
    Counting tasks evaluate a model’s ability to track and count distinct instances or occurrences over time.
    These tasks require precise identification and temporal aggregation of visual elements, which becomes increasingly complex when dealing with fast-paced or dynamic scenarios.
    In extreme sports, for example, counting the number of tricks or jumps requires tracking high-velocity actions and differentiating between overlapping movements in a dynamic, cluttered environment.
\end{itemize}

\subsection{Data Curation Pipeline}
\begin{figure}[tbp]
    \small
    \footnotesize
    \centering
    \includegraphics[width=1.\columnwidth]{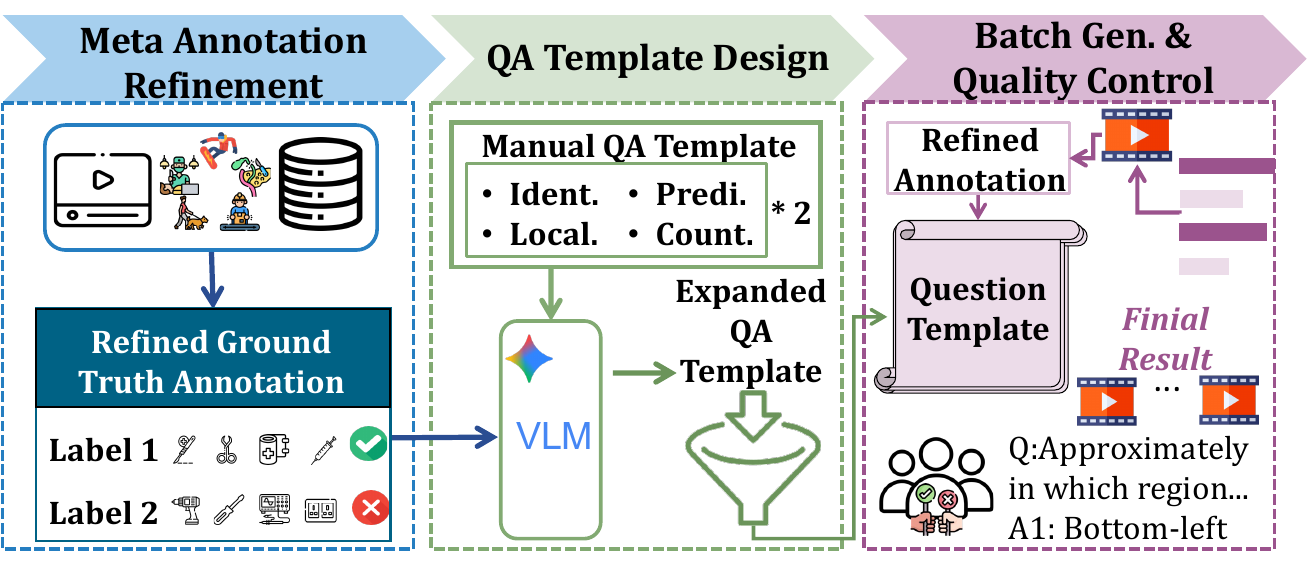}
    \caption{
        Data construction pipeline of EgoCross.
    }
    \label{fig:datacon}
\end{figure}

Based on the selected data sources and question categories, we developed a multi-stage curation pipeline (Fig.~\ref{fig:datacon}) with three key stages: meta annotation refinement, QA template design, and batch generation with quality control. The detailed procedures are described below.

\paragraph{Meta Annotation Refinement.} 
Although the selected datasets provide original annotations, these are typically tailored for simpler, task-specific objectives such as 2D spatial bounding boxes for tool interactions or temporal segments for action classification.
In addition, the annotation formats vary significantly between datasets.
To address this, we performed a comprehensive refinement process that involved unifying annotation formats and conducting manual reviews to ensure the label accuracy.
This refinement step was essential for constructing a reliable ground truth, which serves as the foundation for all subsequent QA generation.

\paragraph{QA Template Design.} 
Following the task taxonomy, we manually designed 8 initial QA templates by creating two for each of the four core task categories. 
To enhance linguistic diversity and complexity, we employed a large language model (Gemini 2.5 pro) to expand the initial templates by generating domain-specific sub-tasks, using the original templates and refined annotations.
All LLM-generated questions were then rigorously verified by human annotators to ensure clarity, logical consistency, and answerability based on the video content.
Annotators also specified the programmatic reasoning steps and the expected answer format (e.g., object names, timestamps) for each template to ensure that the questions are both challenging and solvable.

\paragraph{Batch Generation and Final Quality Control.} 
After obtaining the question templates, we perform batch instantiation to generate final QA pairs.
For each sampled template, we first randomly extract a corresponding video clips based on its predifined duration, and then derive the ground-truth answer by executing the associated programmatic reasoning over the cropped clips. For comprehensive evaluation, we adopt both the traditional closed-form multiple-choice format (CloseQA) and a more flexible open-ended format (OpenQA) for the answers. In CloseQA, each question is accompanied by one correct answer and three distractors randomly sampled from the same answer type. For OpenQA, the answer consists of the full reasoning steps, which is further refined by a LLM.
To ensure the data quality at scale, we conducted a final quality control check by randomly sampling and verifying 10\% of QA pairs from each domain.

\subsection{Dataset Statistics}

Our EgoCross benchmark covers four diverse domains: Surgery, Industry, Extreme Sports (XSports), and Animal Perspective (Animal Per.), sourced from five real-world egocentric video datasets. It comprises 798 video clips and 957 QA pairs, spanning 15 sub-task types grouped into four main categories.
Tab.\ref{tab:dataset_statistics} summarizes key statistics of the five datasets, including the number of clips, QA pairs, and average seconds of video durations (Dur.(s)).
Fig.\ref{fig:task_taxonomy} further illustrates the composition of EgoCross, including: the distribution of the four primary QA task categories, the 15 sub-task types, the question counts across domains, and representative QA examples for each major capability.

\begin{table}[h]
    \centering
    \renewcommand{\arraystretch}{1.2}
    \small
    \footnotesize
     \scalebox{0.85}{
    \begin{tabularx}{\columnwidth}{@{} l | l cc S[table-format=2.1] @{}}
        \toprule
        \textbf{Domain} & \textbf{Source} & \textbf{Clips} & \textbf{QA Pairs} & \textbf{Dur.(s)} \\
        \midrule
        \multirow{2}{*}{Surgery} 
          & CholecTrack20 & 112 & 183 & 29.7 \\
          & EgoSurgery     & 100 & 100 & 20.4 \\
        \midrule
        \multirow{1}{*}{Industry}
          & ENIGMA-51         & 176 & 245 & 16.5 \\
        \multirow{1}{*}{XSports}
          & ExtremeSportFPV   & 242 & 246 & 13.7 \\
        \multirow{1}{*}{Animal Per.}
          & EgoPet         & 168 & 183 & 31.5 \\
        \midrule
        \textbf{EgoCross} & 5 datasets & \textbf{798} & \textbf{957} & \textbf{22.5} \\
        \bottomrule
    \end{tabularx}
    }
    \caption{
        Key statistics of EgoCross benchmark. 
    }
    \label{tab:dataset_statistics}
\end{table}

\section{Experiments}

\begin{table*}[t]
    \centering
    \scalebox{0.95}{
    \begin{tabular}{l|cc|cc|cc|cc|cc}
        \toprule
        \multirow{2}{*}{\textbf{Models}} & \multicolumn{2}{c|}{\thead{\textbf{Surgery}}} & \multicolumn{2}{c|}{\thead{\textbf{Industry}}} & \multicolumn{2}{c|}{\thead{\textbf{XSports}}} & \multicolumn{2}{c|}{\thead{\textbf{Animal Per.}}} & \multicolumn{2}{c}{\thead{\textbf{Overall}}} \\
        & Closed & Open & Closed & Open & Closed & Open & Closed & Open & Closed & Open   \\ 
        \midrule
        \multicolumn{11}{c}{\textit{Proprietary MLLMs}} \\
        \midrule
        \multicolumn{1}{l|}{GPT-4.1} & \underline{57.24} & \underline{39.58} & \textbf{45.71} & 12.24 & \underline{43.09} & \underline{20.33} & \underline{64.48} & \underline{34.43} & \underline{52.63} & \underline{26.65} \\
        \multicolumn{1}{l|}{Gemini 2.5 Pro}  & \textbf{61.48} & \textbf{42.40} & 37.55 & \textbf{24.49} & \textbf{43.90} & \textbf{21.54} & \textbf{68.85} & \textbf{49.18} & \textbf{52.95} & \textbf{34.40} \\
        \midrule
        \multicolumn{11}{c}{\textit{Open-source MLLMs}} \\
        \midrule
        \multicolumn{1}{l|}{Qwen2.5-VL-3B} & 35.69 & 16.96 & 36.33 & 6.94 & 36.59 & 6.91 & 41.53 & 28.42 & 37.54 & 14.81 \\
        \multicolumn{1}{l|}{Qwen2.5-VL-7B} & 46.29 & 21.55 & 37.55 & \underline{22.04} & 41.87 & 6.91 & 53.55 & 31.15 & 44.82 & 20.41 \\
        \multicolumn{1}{l|}{VideoLLaMA3-7B}  & 39.22 & 15.90 & \underline{40.82} & 13.47 & 37.80 & 13.41 & 50.27 & 32.24 & 42.03 & 18.76 \\
        \multicolumn{1}{l|}{InternVL3-8B}  & 47.00 & 17.67 & 33.06 & 11.84 & 41.06 & 11.38 & 49.18 & 30.60 & 42.58 & 17.87 \\
        \midrule
        \multicolumn{11}{c}{\textit{Egocentric MLLMs}} \\
        \midrule
        \multicolumn{1}{l|}{EgoVLPv2}  & 26.50 & - & 34.69 & - & 23.17 & - & 24.04 & - & 27.10 & - \\
        \multicolumn{1}{l|}{EgoGPT} & 31.80 & 13.07 & 24.49 & 10.20 & 24.80 & 13.82 & 41.53 & 26.78 & 30.66 & 15.97 \\
        \bottomrule
    \end{tabular}
    }
    \caption{
        Evaluation results of MLLMs on EgoCross.
        All scores are reported in percentages.
        The best results are marked in \textbf{bold}, and the second-best are \underline{underlined}.
        EgoVLPv2 is not evaluated on open-set tasks due to its model architecture.
    }
    \label{tab:main_results}
\end{table*}

\subsection{Experimental Setup}
\label{sec:setup}

\paragraph{Evaluated Models.}
We select a diverse set of MLLMs spanning three categories to cover major technical paradigms: 1) To assess the current state-of-the-art performance, we include leading proprietary models: GPT-4.1~\cite{achiam2023gpt} and Gemini 2.5 Pro~\cite{comanici2025gemini}. 
2) For open-source general-purpose MLLMs, we consider Qwen2.5-VL (3B, 7B)~\cite{bai2025qwen2}, VideoLLaMA3~\cite{zhang2025videollama}, and InternVL3~\cite{zhu2025internvl3}.
3) To evaluate models tailored for egocentric understanding, we also include two egocentric-specialized models: EgoVLPv2~\cite{pramanick2023egovlpv2}, and EgoGPT~\cite{yang2025egolife}.

\paragraph{Evaluation Metrics.}
Following prior works~\cite{fan2019egovqa, mangalam2023egoschema, plizzari2025omnia}, we use standard \textit{accuracy} metric for CloseQA, which is calculated as the percentage of correctly answered questions.
For OpenQA, we employ a two-stage evaluation process: 1) a direct exact match between the generated and ground-truth answer, and 2) if no match is found, we adopt a \textit{LLM-as-a-Judge} approach to evaluate semantic correctness.
Specifically, Qwen-MAX serves as the judge, providing a binary judgment (Correct/Incorrect) along with a detailed rationale for its decision.
This ensures a robust assessment of semantic equivalence beyond simple string matching.

\paragraph{Implementation Details.}
\label{par:implementation_deatils}
All MLLMs are tested in a zero-shot setting with single-round inference.
For video input, we extract frames at a fixed rate of 0.5 fps.
For datasets that provide pre-sampled frames, we adhere to their original sampling frequency (e.g., EgoSurgery at 0.5 fps and parts of CholecTrack20 at 1 fps).
No maximum frame limit is imposed to allow models to process the full temporal context.
All experiments are conducted on NVIDIA A6000 GPUs.
More implementation details can be found in the Appendix.

\subsection{Results on EgoCross}
\label{sec:main_result}

Evaluation results are summarized in Tab.~\ref{tab:main_results}. 
We analyze the outcomes from four perspectives: 1) task-level challenges, 2) inter-domain variance, 3) model-wise performance, and 4) metric-type analysis.

\noindent\textbf{Task-level Challenges.} 
Most evaluated MLLMs struggle to perform well on our EgoCross benchmark, with average scores falling below 55\% on CloseQA and below 35\% on OpenQA.
Considering that the random guess accuracy for CloseQA is 25\%, these results suggest that the models indeed face substantial challenges in this benchmark.
Furthermore, excluding the top performance from proprietary models (Gemini 2.5 Pro and GPT-4.1), the remaining models perform notably worse, achieving less than 45\% on CloseQA and under 20\% on OpenQA. 
These observations collectively underscore both the difficulty and value of the proposed EgoCross benchmark, while also highlighting the current limitations of state-of-the-art MLLMs in handling cross-domain tasks.

\begin{figure}[tbp]
    \centering
    \begin{subfigure}[b]{0.22\textwidth}
        \centering
        \includegraphics[width=\linewidth]{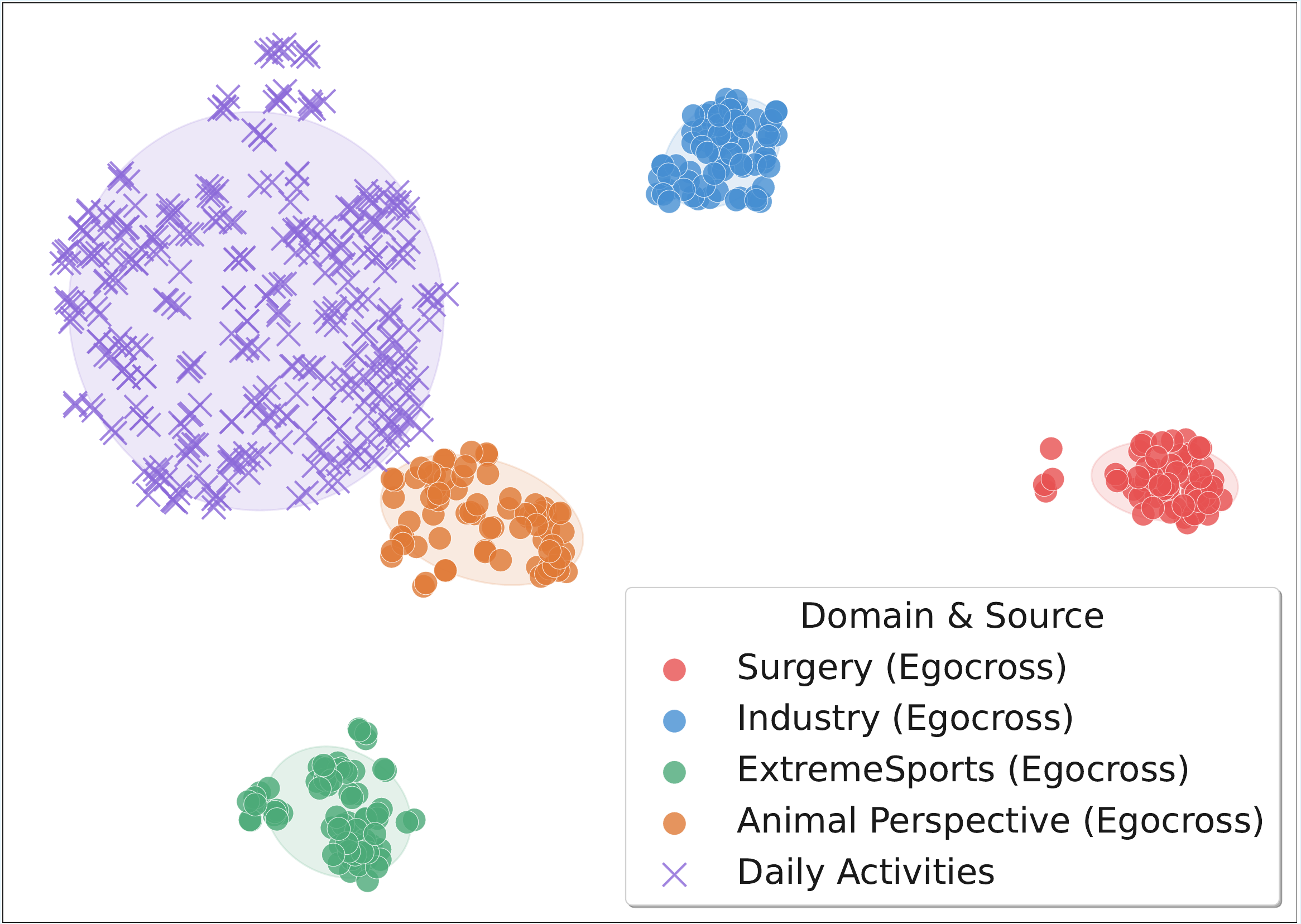}
        \caption{Visual features t-SNE.}
        \label{fig:visual_tsne}
    \end{subfigure}
    \begin{subfigure}[b]{0.22\textwidth}
        \centering
        \includegraphics[width=\linewidth]{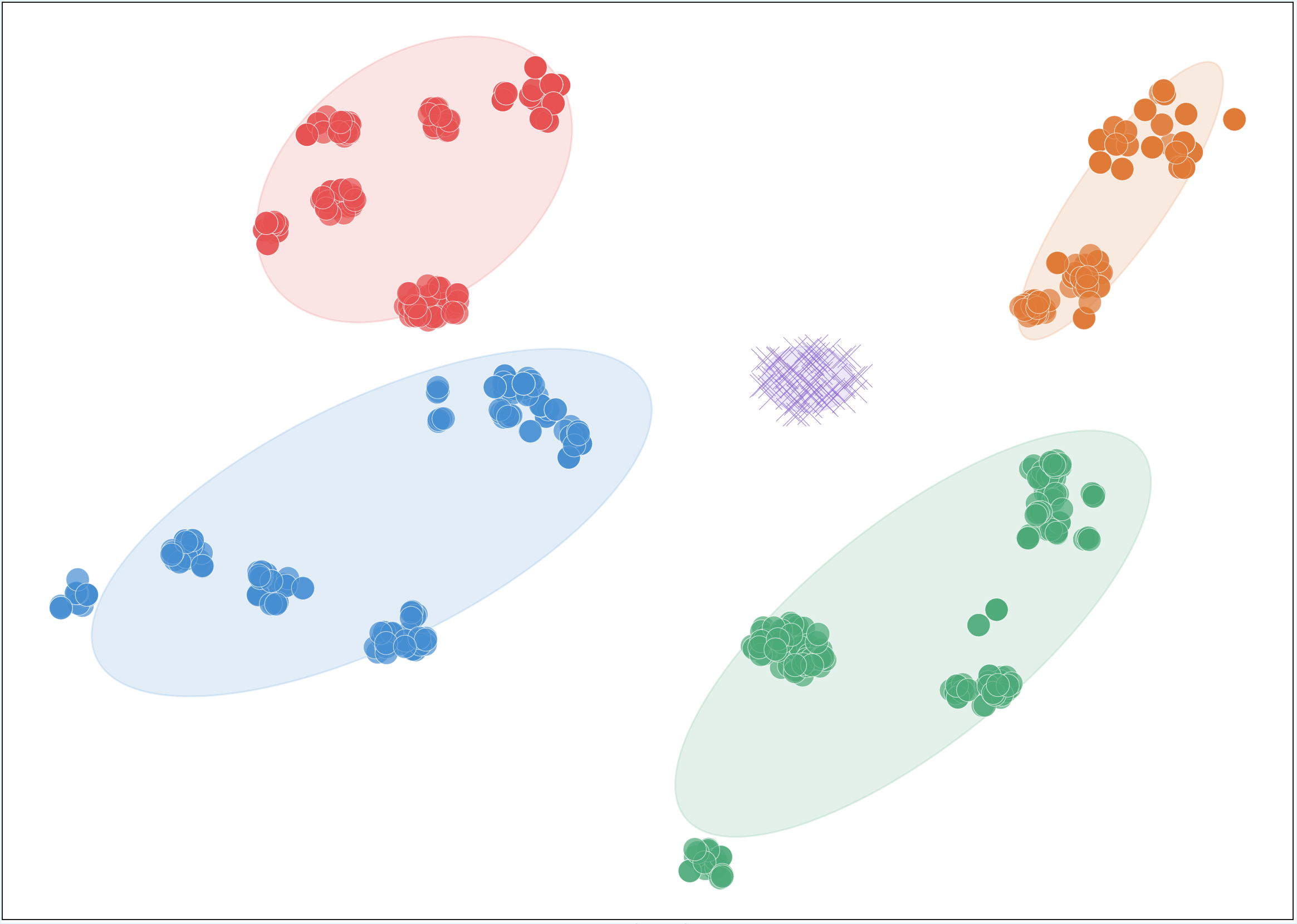}
        \caption{Text features t-SNE.}
        \label{fig:text_tsne}
    \end{subfigure}
    \caption{t-SNE visualization of text and visual features. EgoCross domains are color-coded: Surgery (red), Industry (blue), ExtremeSports (green), Animal Perspective (orange) and Daily-activity (purple).}
    \label{fig:vis}
\end{figure}

\noindent\textbf{Inter-Domain Variance.} 
Across target domains, we observe varying levels of difficulty, ranging from relatively easy (Animal Perspective), middle-hard (Surgery) to particularly challenging (Extreme Sports, Industry). To further investigate inter-domain variance, we visualize t-SNE embeddings of EgoSchema and the four out-of-domain targets, using CLIP~\cite{radford2021learning} as a modality-aligned feature extractor for both visual and textual representations (Fig.~\ref{fig:vis}). The analysis highlights three key findings:
1) All target domains are clearly separated from EgoSchema and from each other in both visual and textual spaces, confirming the dataset’s cross-domain and diverse nature.
2) Within each of our target domains, textual features form sub-clusters, indicating the richness of our QA pairs.
3) The distributions help explain domain difficulty: Animal Perspective is closest to EgoSchema in both modalities, aligning with its relatively easier performance. In contrast, Industry and Extreme Sports are farthest, consistent with their higher difficulty. Surgery seems as a visual outlier but achieves relatively strong performance, suggesting that advanced MLLMs may be more robust to perceptual variation while still challenged by deeper semantic reasoning.

\noindent\textbf{Model-wise Performance.} 
As briefly discussed above, the two proprietary MLLMs achieve the highest overall performance, with Gemini 2.5 Pro outperforming GPT-4.1.
This superiority is expected, given their access to large-scale training data and advanced algorithmic design.
Following them are the open-source models, including Qwen2.5-VL, VideoLLaMA3, and InternVL3.
Compared to proprietary models, they exhibit clear performance drops on both CloseQA and OpenQA tasks, highlighting the substantial gap that remains in tackling cross-domain egocentric understanding within the open-source community.
Surprisingly, the egocentric-specific models (EgoVLPv2, and EgoGPT) perform the worst, despite being explicitly designed and trained on egocentric video data.
Their failure, in contrast to general-purpose models, more clearly underscores the challenge of cross-domain generalization.

\noindent\textbf{Metric-type Analysis.} 
We further analyze the results under different evaluation metrics, namely CloseQA and OpenQA.
Since CloseQA simplifies the task by providing explicit candidate answers, models naturally achieve higher accuracy in CloseQA compared to OpenQA.
Additionally, we observe that CloseQA scores tend to be more stable across different MLLMs, while OpenQA is more sensitive to variations.
For example, GPT-4.1 and Gemini 2.5 Pro achieve nearly identical scores on CloseQA (52.63 vs. 52.95), but differ noticeably on OpenQA (26.65 vs. 34.40).
These observations confirm that OpenQA presents a more challenging setting, reflecting a common limitation of current MLLMs: their generative abilities are generally weaker than their judgment capabilities.
By evaluating both OpenQA and CloseQA, we aim to comprehensively assess the generative and judgmental capacities of MLLMs in cross-domain scenarios.

In addition to dataset-level experiments and analysis, we conduct a more fine-grained evaluation across different QA types.
Due to space limitations, detailed results and discussions are provided in the Appendix.

\begin{figure}[t]
    \centering
    \includegraphics[width=1\columnwidth]{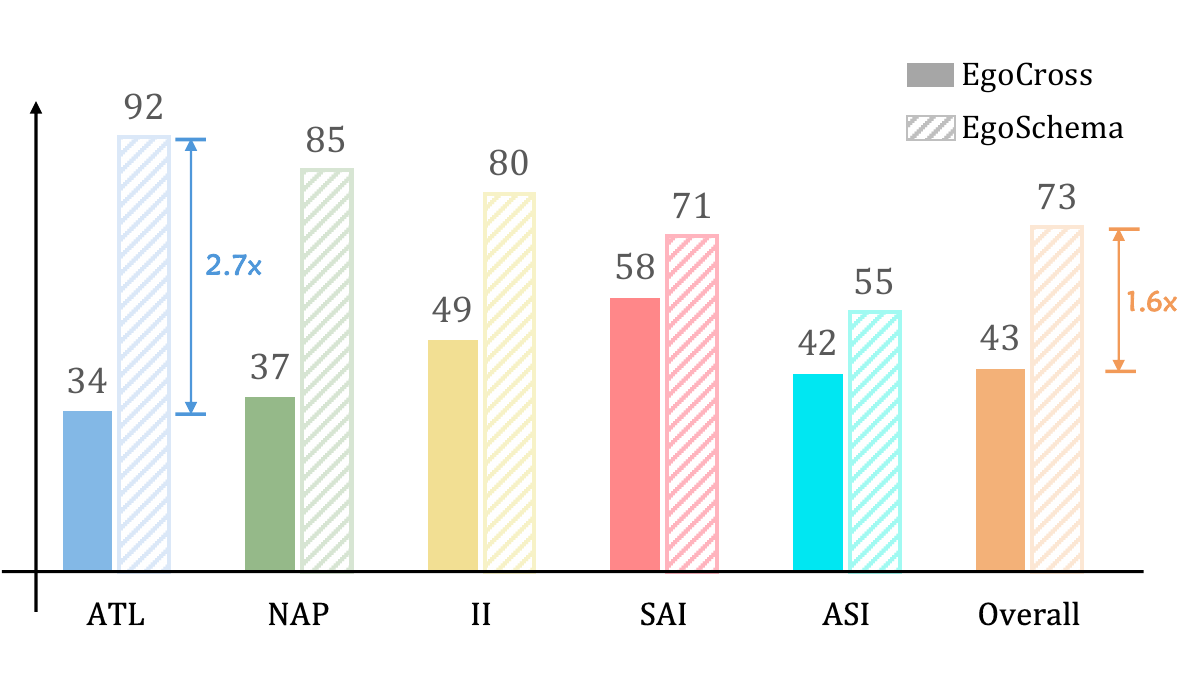}
    \caption{In-domain and cross-domain accuracy comparison across five QA types: Action Temporal Localization (ATL); Next Action Prediction (NAP); Interaction Identification (II); Special Action Identification (SAI); Action Sequence Identification (ASI). The results highlight the performance gap when evaluating on novel domains.}

    \label{fig:cross_domain_comparison}
\end{figure}

\subsection{More Analysis on Cross-Domain Gap}
\label{sec:comparison}

In Sec.~\ref{sec:main_result}, we demonstrate that domain gaps significantly contribute to the overall low performance.
To further investigate this effect and highlight its unique presence in EgoCross, we compare model performance between our benchmark and EgoSchema~\cite{mangalam2023egoschema}, a typical daily-life egocentric dataset featuring common activities like cooking and cleaning. 
To ensure a comparison across analogous QA types, we devise a semi-automated QA type categorization process to align EgoSchema's QA pairs with our tasks. The detailed procedure is provided in the Appendix. Based on these aligned QA pairs, we evaluate Qwen2.5-VL, which achieves the best results among open-source MLLMs, on tasks from both EgoSchema (in-domain) and EgoCross (cross-domain) using the CloseQA protocol.

Results in Fig.~\ref{fig:cross_domain_comparison} reveal a consistent and significant performance drop across all comparable QA types.
For instance, performance on \textit{action temporal localization} drops from an impressive 92.31\% on in-domain EgoSchema to just 34.13\% on the novel domains of surgery, industry, and extreme sports in EgoCross.
Similarly, \textit{next action prediction} accuracy falls from 85.71\% to 37.50\%.
The overall accuracy also drops from 73.58\% to 43.14\%, quantifying the substantial penalty incurred by the domain shift. 
This task-level analysis reinforces our core finding: despite strong results on existing egocentric benchmarks, current MLLMs lack robustness when applied to unseen, domain-specific settings, a limitation largely overlooked in prior work.

\begin{table}[t]
    \centering
    \scalebox{0.8}{
    \begin{tabular}{l|cccc|c}
        \toprule
        \textbf{Method} & \textbf{Surgery} & \textbf{Industry} & \textbf{XSports} & \textbf{Animal Per.} & \textbf{Avg.} \\
        \midrule
        \textcolor{gray}{Baseline*} & \textcolor{gray}{46.29} & \textcolor{gray}{37.55} & \textcolor{gray}{41.87} & \textcolor{gray}{53.55} & \textcolor{gray}{44.82} \\
        Baseline & 37.35 & 35.71 & 34.72 & 43.40 & 37.80 \\
        +Prompt  & 44.58 & 34.29 & 52.78 & 43.40 & 43.76 \\
        +SFT     & 37.35 & 52.86 & 40.28 & 43.40 & 43.47 \\
        +RL      & \textbf{49.40} & \textbf{61.43} & \textbf{54.17} & \textbf{75.47} & \textbf{60.12} \\
        \bottomrule
    \end{tabular}
    }
    \caption{
        CloseQA accuracy in pilot studies.
        “+SFT” and “+RL” denote supervised fine-tuning and reinforcement learning, respectively. * denotes the baseline without vLLM acceleration, it is marked in gray as vLLM acceleration causes slight performance degradation.
    }
    \label{tab:pilot}
\end{table}

\subsection{Pilot Studies}
\label{sec:bridging_gap}

We proactively conduct several pilot studies to explore potential solutions for improving cross-domain egocentric QA.
Specifically, we investigate three techniques: prompt learning, supervised fine-tuning (SFT), and reinforcement learning (RL).
Since both SFT and RL require labeled data, we randomly split the initial test QA pairs into training and testing sets with a 70\%:30\% ratio.
We adopt Qwen2.5-VL-7B as the baseline, and apply vLLM~\cite{kwon2023efficient} for model acceleration.
CloseQA results are shown in Tab.~\ref{tab:pilot}.

The results provide several insights: \textit{1) Overall Trend.} Each method, whether prompting (without labeled data) or SFT/RL (requiring labeled data), improves performance to some extent. This suggests that refining prompt designs, expanding labeled data, and advancing algorithms are all promising directions for future exploration. \textit{2) Impact of SFT.} SFT boosts accuracy in domains like Industry (nearly 20\% improvement). However, in some domains, such as Animal Per. no improvement is observed. This may stem from the inherently higher base performance in this domain, which is closer to natural domains. It could be caused by the limited number of training samples (Animal Per. has only 128 samples, far fewer than other domains.), which restricts the effectiveness of SFT. \textit{3) Effectiveness of RL.} RL shows the most significant improvement across all domains (an average increase of 22\%). We attribute this to RL's ability to learn from a broader range of interactions and feedback during training. The trial-and-error process enables the model to better handle longer sequences and more complex decision-making tasks, allowing it to dynamically adapt to the unique challenges of each domain.

\section{Conclusion}

In this work, we present EgoCross, a new benchmark for evaluating the cross-domain generalization ability of Multimodal Large Language Models (MLLMs) in egocentric video question answering.
EgoCross comprises approximately 1k QA pairs based on video clips carefully collected and curated from four diverse and realistic domains: surgery, industry, extreme sports, and animal perspective, supporting both CloseQA and OpenQA for fine-grained evaluation.
Our extensive evaluation reveals that current SOTA MLLMs struggle to generalize to these unfamiliar domains, despite strong performance on existing benchmarks.
Additionally, we further explore several potential techniques to improve cross-domain generalization.
We believe that EgoCross, together with our experiments and analysis, offers a valuable foundation for future research.


\section*{Acknowledgments}
This work was supported by NSFC grant (No. 62136002 and 62477014), Ministry of Education Research Joint Fund Project (8091B042239), and Fundamental Research Funds for the Central Universities. Project supported by Shanghai Municipal Science and Technology Major Project (2025SHZDZX025G16).

\bibliography{references}

\clearpage
\appendix

\begin{strip}
\centering
{\LARGE \bfseries EgoCross: Benchmarking Multimodal Large Language Models for Cross-Domain Egocentric Video Question Answering\par}
\vspace{0.5em}
{\large \bfseries Supplementary Material\par}
\vspace{1em}
\end{strip}

\input{X_suppl}

\end{document}

%% file: X_suppl.tex

\section{Comparison with Existing Benchmarks}
\label{sec:appendix_benchmark_comparison}

As shown in Table~\ref{tab:benchmark_comparison}, EgoCross sets itself apart from prior benchmarks by uniquely integrating cross-domain challenges, temporal tasks, and a dual Open/Closed-QA format to provide a more rigorous and comprehensive evaluation of model generalization in egocentric video understanding.

\begin{table*}[htbp]
\centering
\renewcommand{\arraystretch}{1.3}
\begin{tabular}{lcccccc}
\toprule
\textbf{Dataset} & \textbf{Cross Domain} & \textbf{Video Length} & \textbf{\# Test} & \textbf{\# Categories} & \textbf{Temporal} & \textbf{QA Types} \\
\midrule
EgoVQA & \ding{55} & (25s, 100s) & 250 & 3 & \ding{55} & OpenQA \\
EgoTaskQA & \ding{55} & 25s & 8k & 4 & \ding{55} & OpenQA \\
EgoSchema & \ding{55} & 3 min & 500 & - & \ding{55} & CloseQA \\
EgoThink & \ding{55} & - & 750 & 12 & \ding{55} & OpenQA \\
EgoTempo & \ding{55} & 45s & 500 & 10 & \ding{51} & OpenQA \\
\midrule
\textbf{EgoCross} & \ding{51} & 22.5s & 957 & 15 & \ding{51} & CloseQA \& OpenQA \\
\bottomrule
\end{tabular}
\caption{Overview of each dataset’s characteristics, including average video length, number of test examples, number of categories, number of scenes where videos are captured, and question types (OpenQA or CloseQA).}
\label{tab:benchmark_comparison}
\end{table*}

\section{More Implement Details}

This section provides a more detailed description of our experimental setup to support reproducibility and enable fair comparisons.

\subsection{Implement Details of Main Experiments}
This section supplements the settings from Section~\ref{sec:setup} to ensure full reproducibility of results reported in Section~\ref{sec:main_result} and a fair comparison across all models.

Our evaluation methodology is based on a zero-shot, single-round inference paradigm, executed on NVIDIA A6000 GPUs. As previously mentioned, video frames are sampled at dataset-specific rates (0.5-1 fps) with no maximum frame limit. To guarantee deterministic and reproducible outputs, we set the key inference parameter \texttt{do\_sample=False} for greedy decoding across all experiments.
The exact prompts used for each task are detailed below to ensure a fair and standardized evaluation.

\begin{figure}[ht]
    \centering
    \includegraphics[width=1\linewidth]{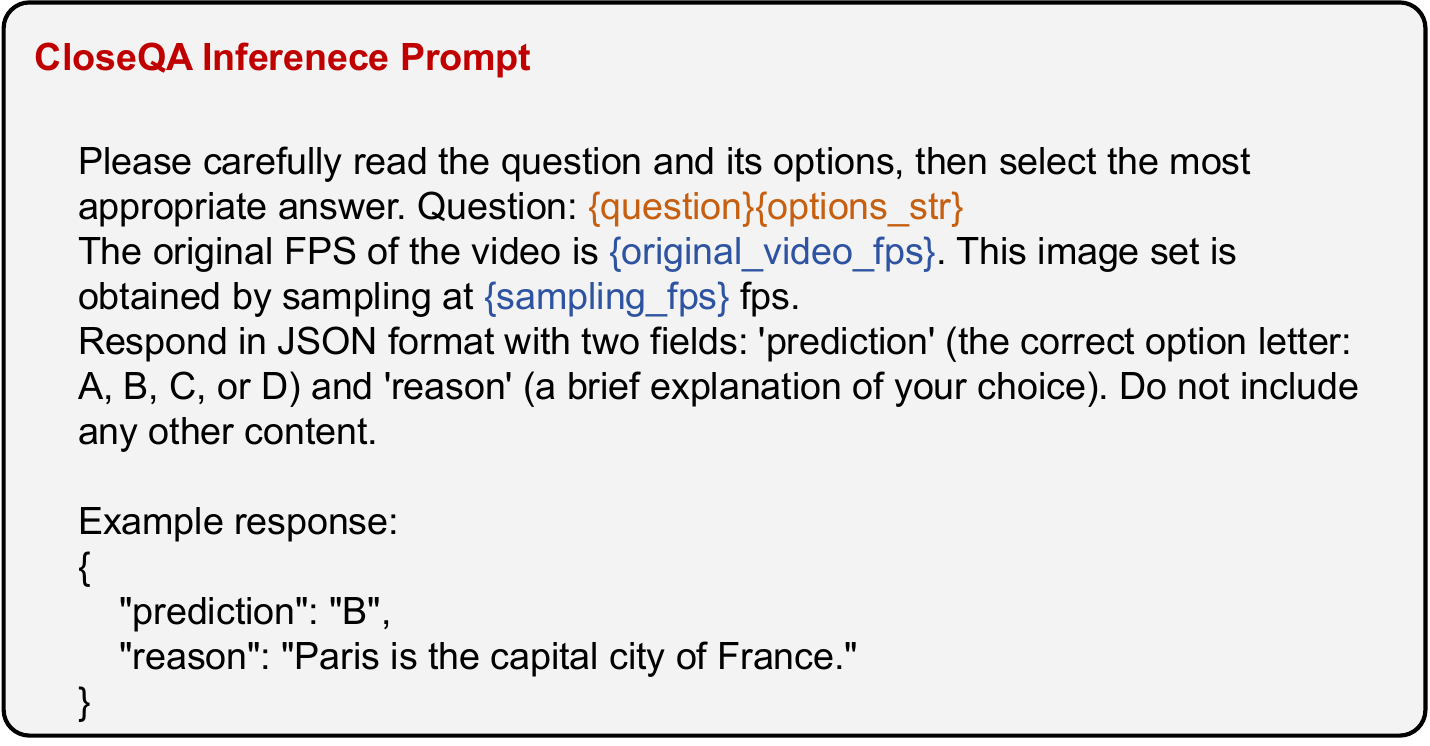}
    \caption{Prompt of CloseQA Inferenece}
    \label{fig:closeqa_prompt}
\end{figure}

\paragraph{Close-ended Question Answering (CloseQA).}
For multiple-choice questions, the prompt illustrated in Figure~\ref{fig:closeqa_prompt} instructs the model to return its answer and reasoning in a structured JSON format. The letter following the ``prediction'' is then extracted as the final answer.

\begin{figure}[ht]
    \centering
    \includegraphics[width=1\linewidth]{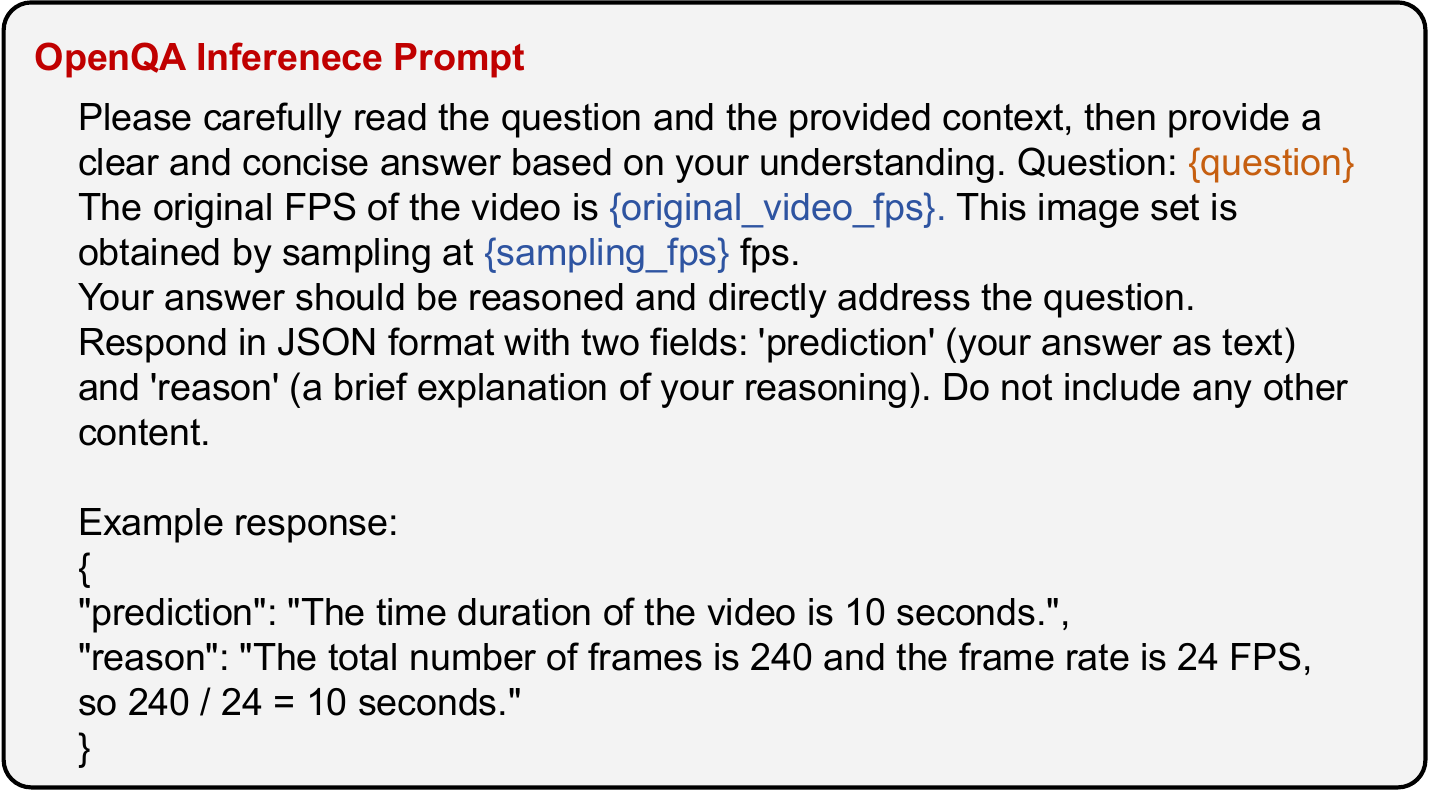}
    \caption{Prompt of OpenQA Inferenece}
    \label{fig:openqa_prompt}
\end{figure}

\paragraph{Open-ended Question Answering (OpenQA).}
For free-form questions, the prompt illustrated in Figure~\ref{fig:openqa_prompt} guides the model in generating a textual answer. The string after the
``prediction'' is used as the model response for evaluation.

\begin{figure}[ht]
    \centering
    \includegraphics[width=1\linewidth]{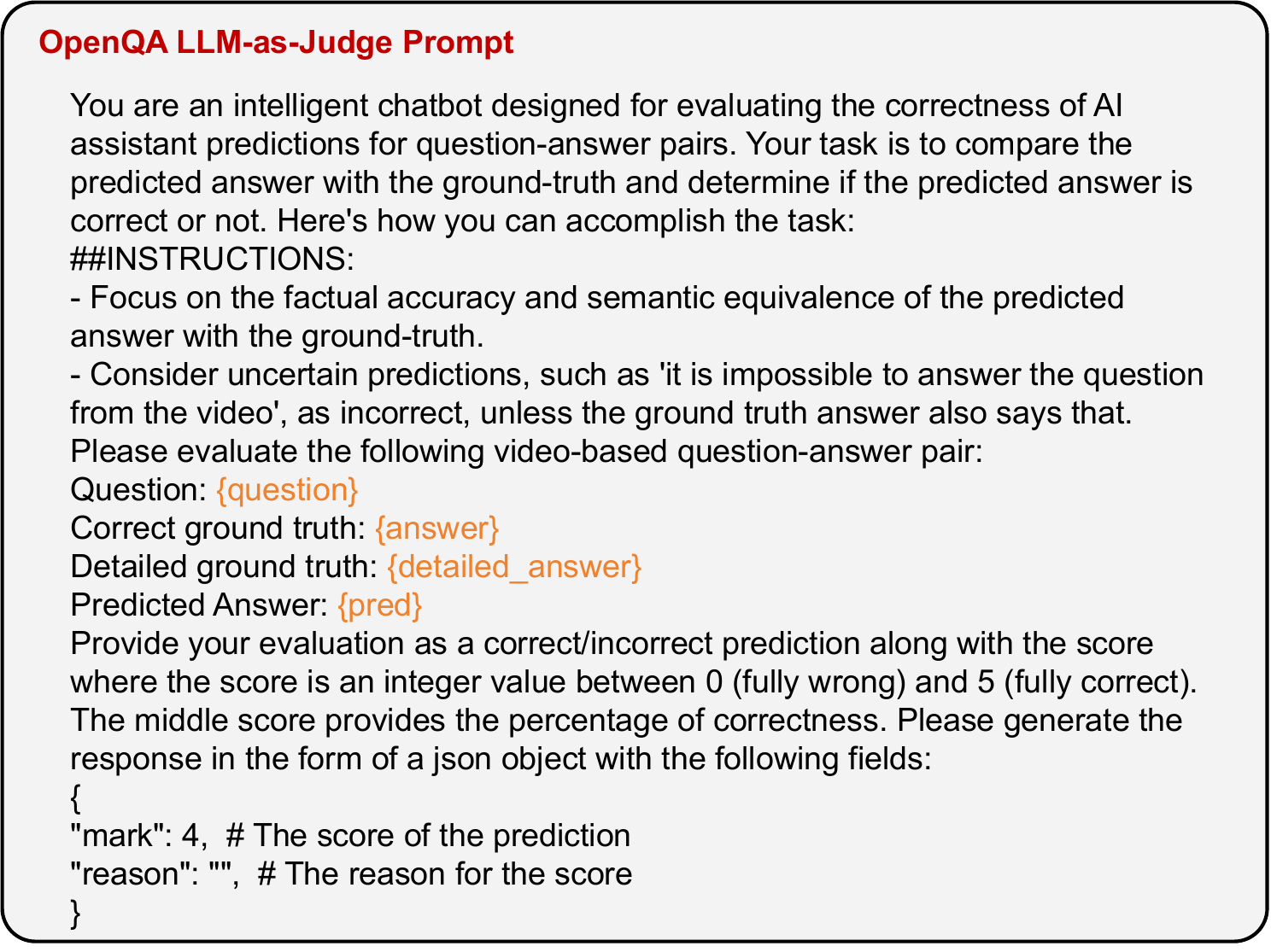}
    \caption{Prompt of LLM-as-Judge}
    \label{fig:llm_judge_prompt}
\end{figure}

\paragraph{Evaluation Protocol for OpenQA.}
Given the subjective nature of open-ended answers, we employ Qwen-Max as an automated judge to ensure a consistent and scalable evaluation. The LLM judge evaluates the semantic correctness of the model prediction against the ground truth ground truth based on the prompts as shown in Figure~\ref{fig:llm_judge_prompt}

\subsection{Implement Details of Domain Gap Experiments}
\label{sec:appendix_egoschema_categorization}

As mentioned in Section~\ref{sec:comparison}, to enable a direct comparison between the performance on EgoSchema (in-domain) and EgoCross (cross-domain), it was necessary to align the QA pairs from EgoSchema with our predefined task categories. Since EgoSchema does not provide official task-type labels, we developed a semi-automated, iterative categorization process. This section details the methodology, the classification rules, and the final distribution of the aligned QA pairs.

\subsubsection{Methodology for Semi-Automated Categorization}

To align the questions in EgoSchema with our predefined task categories, we employed a semi-automated workflow. we prompted a large language model (LLM) to generate keywords for each templates of task type in EgoCross. These keywords were extracted to capture the core concepts and characteristics of each task type, such as identification, localization, prediction, and counting. We then used these keyword sets to classify the questions in EgoSchema by matching them to the relevant keywords. For categories with a limited number of questions, we avoided further subdivision to prevent statistical instability.

Crucially, this initial classification was refined through a rigorous human-in-the-loop process. Human experts reviewed the automated results, analyzed the question and options of misclassified instances, and iteratively updated the keyword lists. This refinement cycle was repeated for five rounds until the categorization stabilized and achieved high accuracy, as confirmed by final manual validation. This methodology produced a refined and consistent set of questions fully aligned with our task definitions.

\subsubsection{Classification Rules and Keywords}
The refinement process yielded a set of keywords and a hierarchy to resolve ambiguities.

\paragraph{Classification Hierarchy.} A strict priority order was established for overlapping cases: \\
\textbf{1) Main Categories:} Prediction $\rightarrow$ Counting $\rightarrow$ Localization $\rightarrow$ Identification. \\
\textbf{2) Identification Subtypes:} Dominant Held Object $\rightarrow$ Action Sequence $\rightarrow$ Interaction $\rightarrow$ Special Action.

\paragraph{Final Keyword Sets.}
The final keyword sets and question distribution are detailed in Table~\ref{tab:keyword_sets_distribution}. To handle questions unique to EgoSchema, we designated \texttt{Inference Prediction} (which is absent in EgoCross) and created \texttt{Action State Identification} as a catch-all category for general action identification queries. The remaining six sub-tasks with non-zero counts constitute the core set of aligned task types used for our direct cross-dataset performance comparison.

\paragraph{Quantitative Cross-Domain Comparison.}
\label{sec:appendix_cross_domain_quant}
The significant performance gap between the two domains, illustrated in Figure~\ref{fig:cross_domain_comparison}, is quantitatively detailed in Table~\ref{tab:cross_domain_comparison}. To measure this gap, we evaluated Qwen2.5-VL on both the full datasets and our carefully aligned subsets.
On the aligned subset of questions, the model's accuracy plummets from \textit{73.58\%} on in-domain EgoSchema tasks to just \textit{43.14\%} on their cross-domain EgoCross counterparts. Notably, this performance trend on the subset closely mirrors the model's overall results on the full datasets (\textit{69.60\%} on EgoSchema vs. \textit{44.31\%} on EgoCross), validating that our aligned subset serves as a reliable proxy for evaluating the domain gap. The table further breaks down this degradation across analogous task types, highlighting that the domain shift poses a significant and consistent challenge to models' capabilities.

\begin{table}[!h]
\centering
\small 

\begin{tabular}{lcccc}
\toprule
& \multicolumn{2}{c}{\textbf{In-Domain}} & \multicolumn{2}{c}{\textbf{Cross-Domain}} \\
\cmidrule(lr){2-3} \cmidrule(lr){4-5}
\textbf{Question Type} & {\textbf{Num.}} & {\textbf{Acc. (\%)}} & {\textbf{Num.}} & {\textbf{Acc. (\%)}} \\ \midrule
Special Action Id. & 56 & 71.43 & 46 & 58.70 \\
Interaction Id. & 50 & 80.00 & 104 & 49.04 \\
Temporal Loc. & 26 & 92.31 & 126 & 34.13 \\
Action Sequence Id. & 47 & 55.32 & 50 & 42.00 \\
Next Action Pred. & 14 & 85.71 & 24 & 37.50 \\
\midrule
\textbf{Overall Above} & \textbf{193} & \textbf{73.58} & \textbf{350} & \textbf{43.14} \\
\textbf{Overall Dataset} & \textbf{500} & \textbf{69.60} & \textbf{957} & \textbf{44.31} \\
\bottomrule
\end{tabular}
\caption{A direct comparison of Qwen2.5-VL's performance on analogous task types between the in-domain EgoSchema benchmark and our cross-domain EgoCross Benchmark. This highlights the stark performance degradation when transitioning to novel domains, even on structurally similar tasks.}
\label{tab:cross_domain_comparison}
\end{table}

\begin{table*}[t]
    \centering
    \begin{tabularx}{0.9\textwidth}{@{} l l l X @{}} 
        \toprule
        \textbf{Category} & \textbf{Sub-tasks} & \textbf{Count} & \textbf{Keywords} \\
        \midrule
        \multirow{4}{*}{Prediction} & Next Action Prediction & 14 & \textit{what will happen, future action, next phase, next direction, prepare for, ready for} \\
        \addlinespace
         & Inference Prediction & 26 & \textit{taking into account, analyze, evaluate, compare, discuss, deduce, overall focus} \\
        \addlinespace
        \hline
        \addlinespace
        Counting & Object Counting & 0 & \textit{how many, number of, count how, quantity, total number} \\
        \addlinespace
        \hline
        \addlinespace
        \multirow{4}{*}{Localization} & Action Temporal Localization & 26 & \textit{at what time, when did start/end/occur, key moments when, before, after, during} \\
        \addlinespace
         & Object Spatial Localization & 0 & \textit{where is/are, in which region/location, where located} \\
        \addlinespace
        \hline
        \addlinespace
        \multirow{12}{*}{Identification} & Dominant Held-Object Identification & 10 & \textit{primary/main tool, tool used, effectiveness tools, how tools contribute} \\
        \addlinespace 
         & Action Sequence Identification & 47 & \textit{sequence of actions, from start to finish, key steps, main stages, overarching process} \\
        \addlinespace
         & Interaction Identification & 50 & \textit{interaction between, two characters, both characters, collaborate, relationship} \\
        \addlinespace
         & Special Action Identification & 56 & \textit{most significant/important/critical, key turning points, pivotal, vital steps} \\
        \addlinespace
         & Sport/Animal/Not Visible Identification & 0 & \textit{sport, game play, animal, pet, not visible/shown/present} \\
        \addlinespace
         & Action State Identification & 239 & \textit{primary objective/goal, describe, summarize, explain, infer, deduce, what is/was primary} \\
        \bottomrule
    \end{tabularx}
    \caption{Final Keyword Sets and Question Distribution for EgoSchema Task Categorization.}
    \label{tab:keyword_sets_distribution}
\end{table*}

\subsection{Implement Details Of Pilot Studies}
\label{sec:appendix_pilot_details}

\paragraph{General Setup.}
To ensure consistency and efficiency across experiments, we employ the vLLM framework~\cite{kwon2023efficient} for both training and inference with the Qwen2.5-VL-7B model. While vLLM significantly accelerates these processes, we note a slight performance trade-off as shown in Table~\ref{tab:pilot}, which is an acceptable compromise, particularly for the extensive sampling required in RL training. All experiments were conducted on a server equipped with 8 NVIDIA H100 GPUs. The training setups for SFT and RL are inspired by the Video-R1 project, and we plan to release our code for reproducibility. For video inputs during training, we sample between 4 to 16 frames; sequences longer than 16 frames are truncated to 16. The inference settings for the pilot studies are kept consistent with those used for the main results.

\begin{figure}[h]
    \centering
    \includegraphics[width=.9\linewidth]{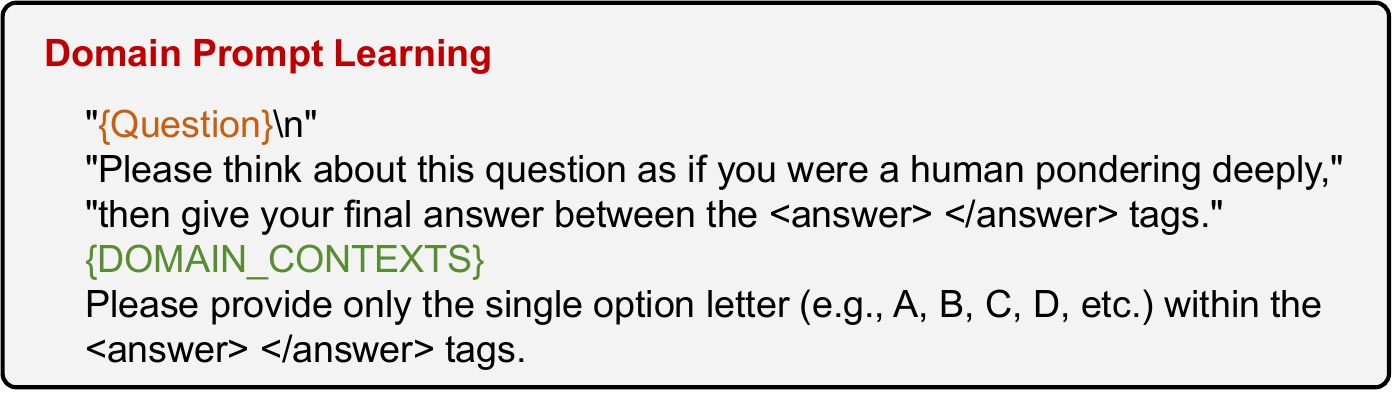}
    \vspace{0.2cm} 
    \includegraphics[width=.9\linewidth]{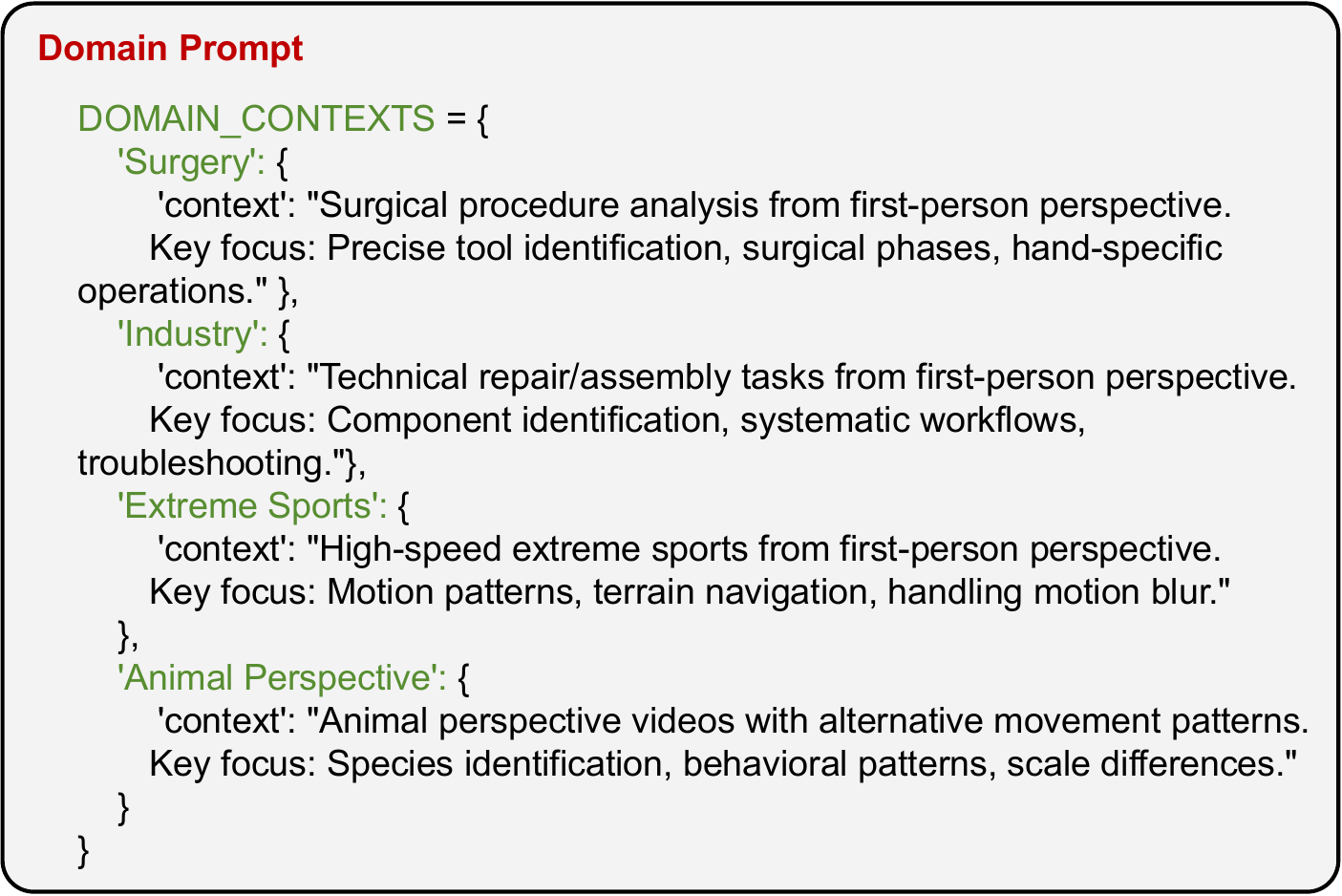}
    \caption{The two-part prompt used for domain-specific prompt learning. The first part (top) provides domain context, and the second part (bottom) presents the specific question.}
    \label{fig:domain_prompt}
\end{figure}

\paragraph{Prompt Learning.}
The prompt for this method consists of two parts. The first part provides domain-specific context and examples, while the second part poses the direct question about the input video. This structure, illustrated in Figure~\ref{fig:domain_prompt}, aims to guide the model towards the specific characteristics of each domain before it attempts to answer the question.

\begin{figure}[h]
    \centering
    \includegraphics[width=1\linewidth]{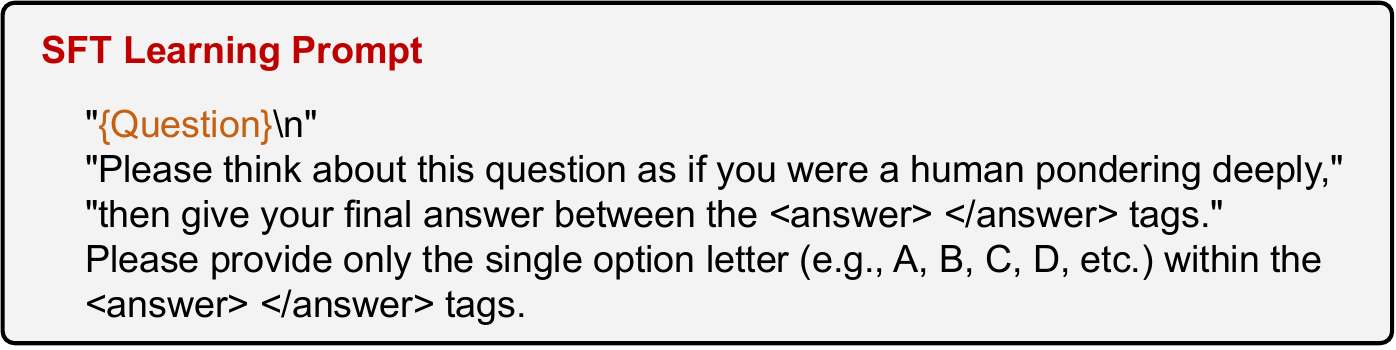}
    \caption{The data format for SFT. Each sample consists of a video and a conversation, where the model is trained to generate the assistant's response based on the user's query.}
    \label{fig:sft_prompt}
\end{figure}

\paragraph{Supervised Fine-tuning (SFT).}
We perform full-parameter SFT on the base model, updating all of its weights. The training utilizes 4 H100 GPUs, with a per-device batch size of 1 and 2 gradient accumulation steps, resulting in an effective batch size of 8. We set the learning rate to \textit{1e-6} and train for 12 epochs. To manage memory and accelerate training, we leverage a suite of optimizations including DeepSpeed ZeRO-2, BF16 mixed precision, gradient checkpointing, and Flash Attention 2. The data format used for SFT is shown in Figure~\ref{fig:sft_prompt}.

\begin{figure}[tbp]
    \centering
    \includegraphics[width=1\linewidth]{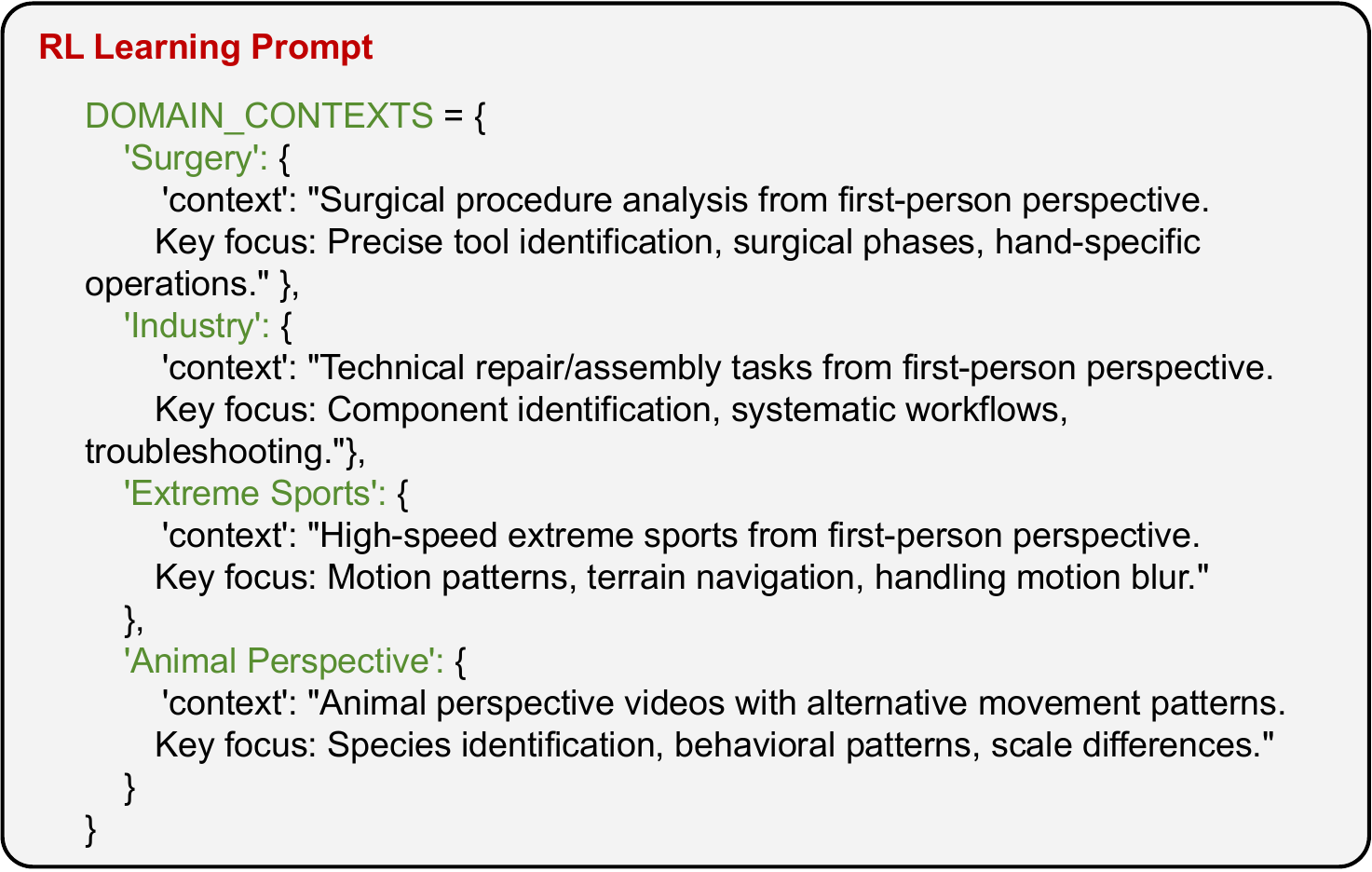}
    \caption{The prompt structure for RL training. The model generates a response, which is then evaluated by a reward model to provide feedback for policy optimization.}
    \label{fig:rl_prompt}
\end{figure}

\paragraph{Reinforcement Learning (RL).}
Our RL approach, based on Generative Reward-based Policy Optimization (GRPO), trains the model from scratch without an SFT warm-up. The training is distributed across all 8 H100 GPUs. We use a learning rate of \textit{1e-6} with a cosine scheduler and train for 16 epochs. The optimization strategy is intensified with DeepSpeed ZeRO-3 to accommodate the RL process, alongside BF16, gradient checkpointing, and Flash Attention 2. Key RL-specific hyperparameters include a reward-shaping beta of 0.04 and generating 8 responses per prompt during training for policy updates. The prompt structure for RL is depicted in Figure~\ref{fig:rl_prompt}.

\section{More Experiment Results}
\label{sec:appendix_detailed_results}

\subsection{CloseQA and OpenQA evaluations}
This section provides detailed results for both CloseQA and OpenQA evaluations, broken down by task capabilities across different domains. The results are presented in Table~\ref{tab:detailed_results_closeset} and Table~\ref{tab:detailed_results_openset}.

To dissect the sources of this performance degradation, Table~\ref{tab:detailed_results_closeset} and Table~\ref{tab:detailed_results_openset} provide a fine-grained analysis of CloseQA and OpenQA capabilities, respectively. This detailed breakdown reveals that the generalization gap is not uniform across different skills. In both evaluation settings, the proprietary SOTA models demonstrate a clear superiority in tasks requiring complex reasoning. This is most evident in the OpenQA \textbf{Prediction (P)} capability (Table~\ref{tab:detailed_results_openset}), where Gemini 2.5 Pro achieves a remarkable \textit{62.50\%} on surgical prediction, a task where most other models score in the single digits. This highlights that advanced temporal and causal reasoning is a key differentiator. In contrast, the performance gap narrows for more direct perceptual tasks like \textbf{Counting (C)}, where several open-source models exhibit competitive performance. This detailed view also uncovers intriguing failure modes. For instance, in the Animal Perspectives-Localization task, Gemini 2.5 Pro achieves a high score of \textit{42.42\%}, while GPT-4.1 unexpectedly scores zero. A qualitative review reveals that while Gemini correctly provides timestamps, GPT-4.1 defaults to referencing frame indices (e.g., "The cat first interacts with the plastic in the fourth image"), failing to follow the prompt's explicit instruction to use the provided video FPS. This specific failure, not observed in other domains for the same model, suggests its instruction-following capability can be brittle and context-dependent. Ultimately, this granular analysis confirms the primary bottleneck: the challenge lies less in basic perception and more in the robust application of high-level skills—such as temporal reasoning, knowledge integration, and consistent instruction-following—within novel, specialized contexts.

\begin{table*}[htbp]
\centering
\normalsize
\setlength{\tabcolsep}{3.5pt}

\sisetup{table-format=2.2} 
\begin{tabular}{l|cccc|cccc|ccc|cc}
\toprule
\multirow{2}{*}{\textbf{Models}} & \multicolumn{4}{c|}{\textbf{Surgery}} & \multicolumn{4}{c|}{\textbf{Industry}} & \multicolumn{3}{c|}{\textbf{XSports}} & \multicolumn{2}{c}{\textbf{\thead{Animal Per.}}} \\
\cmidrule(lr){2-5} \cmidrule(lr){6-9} \cmidrule(lr){10-12} \cmidrule(lr){13-14}
& C & I & L & P & C & I & L & P & I & L & P & I & L \\
\midrule
\multicolumn{14}{c}{\textit{Open-source MLLMs}} \\
\midrule
Qwen2.5-VL-3B   & 34.00 & 44.34 & 30.38 & 27.08 & 29.69 & \underline{40.62} & 34.38 & 41.51 & 53.15 & 25.33 & 20.00 & 47.01 & 31.82 \\
Qwen2.5-VL-7B   & \underline{58.00} & 50.00 & \underline{39.24} & 37.50 & 29.69 & 37.50 & 45.31 & 37.74 & 52.25 & \textbf{29.33} & \textbf{38.33} & 52.99 & \textbf{54.55} \\
InternVL3-8B    & \textbf{66.00} & 54.72 & 35.44 & 29.17 & 26.56 & 29.69 & 34.38 & \underline{43.40} & 58.56 & \underline{28.00} & 25.00 & 53.85 & 40.91 \\
VideoLLaMA3-8B  & 34.00 & 46.23 & \underline{39.24} & 29.17 & \underline{40.62} & \underline{40.62} & 26.56 & \textbf{58.49} & 52.25 & 22.67 & \underline{30.00} & 60.68 & 31.82 \\
\midrule
\multicolumn{14}{c}{\textit{Egocentric MLLMs}} \\
\midrule

EgoGPT          & 36.00 & 36.79 & 24.05 & 29.17 & 12.50 & 21.88 & 28.12 & 37.74 & 29.73 & 22.67 & 18.33 & 47.86 & 30.30 \\
EgoVLPv2        & 20.00 & 33.02 & 26.58 & 18.75 & \textbf{45.31} & 29.69 & 31.25 & 32.08 & 21.62 & 24.00 & 25.00 & 25.64 & 21.21 \\
\midrule
\multicolumn{14}{c}{\textit{Proprietary MLLMs}} \\
\midrule
GPT-4.1         & 54.00 & \underline{68.87} & 35.44 & \underline{70.83} & 34.38 & \textbf{54.69} & \underline{51.56} & 41.51 & \textbf{66.67} & 26.67 & 20.00 & \textbf{79.49} & 37.88 \\
Gemini 2.5 Pro  & 34.00 & \textbf{70.75} & \textbf{55.70} & \textbf{79.17} & 15.62 & 39.06 & \textbf{64.06} & 30.19 & \underline{63.06} & \textbf{29.33} & 26.67 & \underline{78.63} & \underline{51.52} \\
\bottomrule
\end{tabular}
\caption{Evaluation on Close-ended Questions. The task types are Counting (C), Identification (I), Localization (L), and Prediction (P). Best results are marked in \textbf{bold}, and the second-best is \underline{underlined}.}
\label{tab:detailed_results_closeset}
\end{table*}

\begin{table*}[htbp]
\centering
\normalsize
\setlength{\tabcolsep}{3.5pt}
\sisetup{table-format=2.2} 

\begin{tabular}{l|cccc|cccc|ccc|cc}
\toprule
\multirow{2}{*}{\textbf{Models}} & \multicolumn{4}{c|}{\textbf{Surgery}} & \multicolumn{4}{c|}{\textbf{Industry}} & \multicolumn{3}{c|}{\textbf{Xsports}} & \multicolumn{2}{c}{\textbf{Animal Per.}} \\
\cmidrule(lr){2-5} \cmidrule(lr){6-9} \cmidrule(lr){10-12} \cmidrule(lr){13-14}
& C & I & L & P & C & I & L & P & I & L & P & I & L \\
\midrule
\multicolumn{14}{c}{\textit{Open-source MLLMs}} \\
\midrule
Qwen2.5-VL-3B   & \textbf{50.00} &  6.60 & 13.92 & 10.42 &  0.00 & 14.06 & 12.50 &  0.00 &  6.31 &  2.67 & 13.33 & 42.74 &  3.03 \\
Qwen2.5-VL-7B   & \textbf{50.00} &  6.60 & 31.65 &  8.33 & 18.75 & \textbf{45.31} & 18.75 & \underline{1.89} &  9.01 &  1.33 & 10.00 & 43.59 &  9.09 \\
InternVL3-8B    & 40.00 & 15.09 & 12.66 &  8.33 &  1.56 & \underline{25.00} & 18.75 &  0.00 &  8.11 & 12.00 & 16.67 & 45.30 &  4.55 \\
VideoLLaMA3-8B  & \underline{46.00} &  6.60 & 17.72 &  2.08 & \textbf{28.12} & 10.94 & 10.94 & \underline{1.89} &  8.11 & \underline{21.33} & 13.33 & 48.72 &  3.03 \\
\midrule
\multicolumn{14}{c}{\textit{Egocentric MLLMs}} \\
\midrule
EgoGPT          & 38.00 &  4.72 & 13.92 &  4.17 &  4.69 & 12.50 & \underline{21.88} &  0.00 &  6.31 & \underline{21.33} & 18.33 & 41.88 &  0.00 \\
\midrule
\multicolumn{14}{c}{\textit{Proprietary MLLMs}} \\
\midrule
GPT-4.1         & \underline{46.00} & 31.13 & 40.51 & \underline{50.00} &  9.38 & 18.75 & 15.62 & \textbf{3.77} & \textbf{21.62} &  6.67 & 35.00 & \textbf{53.85} &  0.00 \\
Gemini 2.5 Pro  & 36.00 & \textbf{33.96} & \textbf{45.57} & \textbf{62.50} & \underline{26.56} & \underline{25.00} & \textbf{40.62} & \underline{1.89} & \underline{17.12} & 14.67 & \textbf{38.33} & \underline{52.99} & \textbf{42.42} \\
\bottomrule
\end{tabular}
\caption{Detailed evaluation on Open-ended Questions, broken down by capability type: Counting (C), Identification (I), Localization (L), and Prediction (P). All scores are in percentages. Best results are marked in \textbf{bold}, and the second-best is \underline{underlined}.}
\label{tab:detailed_results_openset}
\end{table*}

\section{More Details of Datasets}
\label{sec:appendix_subtasks}
\begin{figure*}[!htbp]
    \centering
    \includegraphics[width=\textwidth]{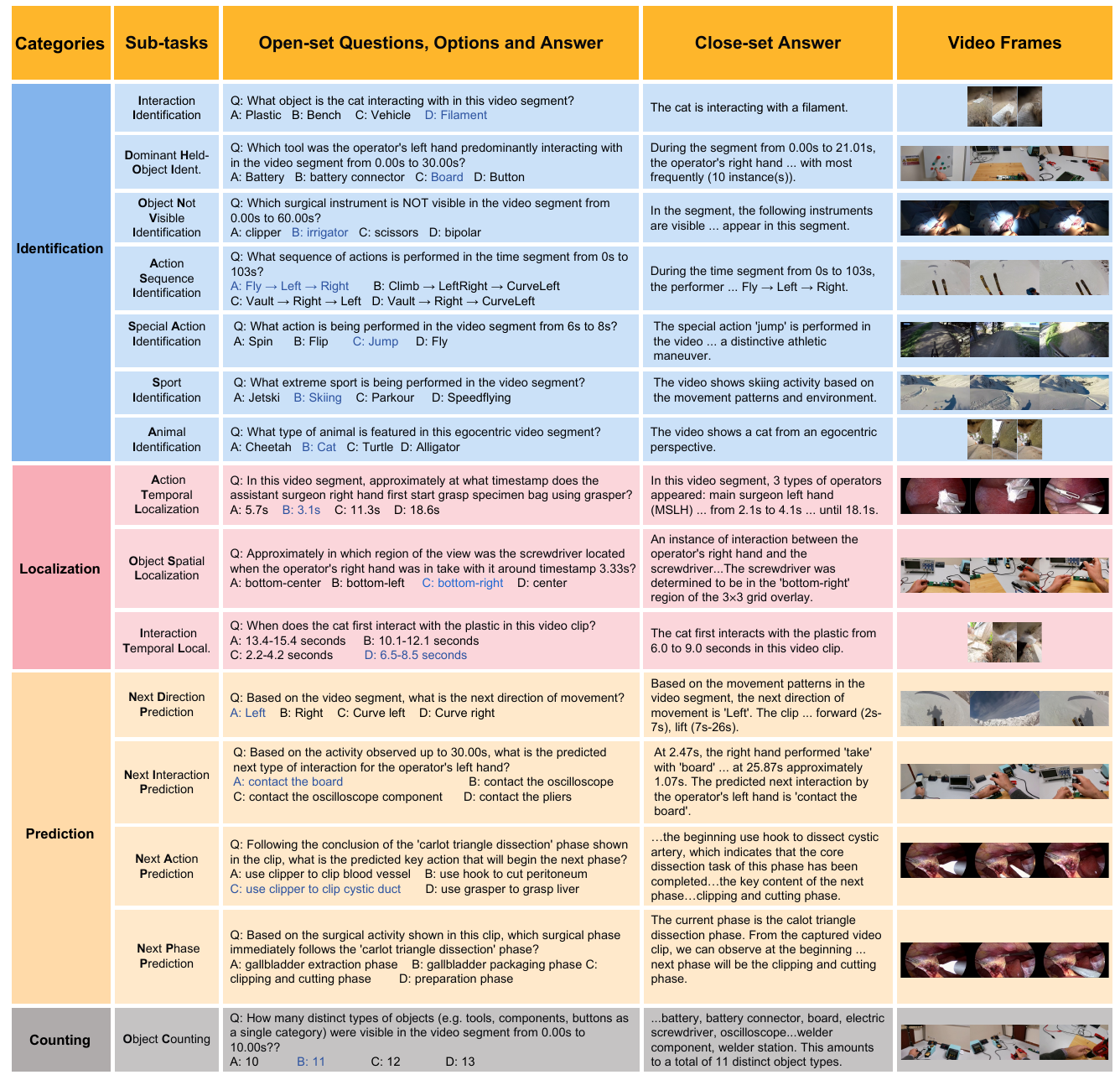} 
    \caption{
        \textbf{Detailed examples of the 15 sub-tasks across our four core categories: Identification, Localization, Prediction, and Counting.} For each sub-task, we present a representative question, the corresponding answer, and a visual frame from the video. These examples illustrate the diversity of our benchmark, spanning from low-level perception (\textit{e.g.}, object identification, spatial localization) to high-level reasoning (\textit{e.g.}, next action prediction, counting dynamic events).
    }
    \label{fig:appendix_subtasks_examples}
\end{figure*}

This appendix provides detailed illustrative examples for the 15 sub-tasks within our proposed QA task taxonomy. As mentioned in the main text, these tasks are grouped into four core categories: \textbf{Identification}, \textbf{Localization}, \textbf{Prediction}, and \textbf{Counting}. Figure~\ref{fig:appendix_subtasks_examples} showcases a representative question, closed answer, open answer and three corresponding visual frames for each sub-task, highlighting the diverse challenges across different egocentric video domains.

These examples cover scenarios ranging from fine-grained object recognition in surgical footage to high-level action anticipation in dynamic outdoor scenes, reflecting the breadth of reasoning skills required. Each sub-task is carefully crafted to target specific dimensions of egocentric understanding, such as detecting subtle hand–object interactions and predicting forthcoming activities from partial observations. Collectively, they constitute a rigorous and comprehensive benchmark for assessing multimodal models in realistic egocentric settings, where successful performance depends on the seamless integration of temporal context, visual detail, and task knowledge.

\begin{figure*}[!htbp]
    \centering
    \includegraphics[width=\textwidth, height=\textheight, keepaspectratio]{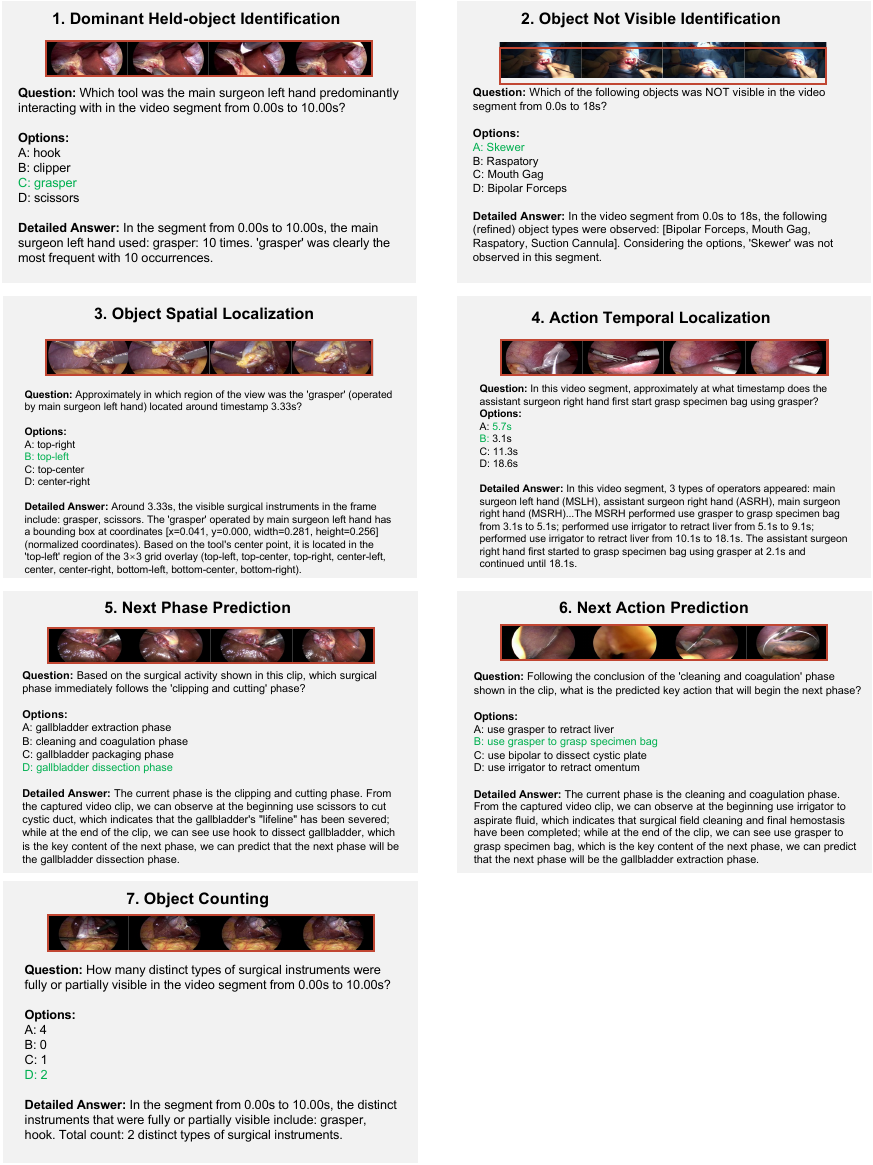} 
    \caption{Representative QA examples from the Surgery domain.}
    \label{fig:appendix_surgery}
\end{figure*}

\begin{figure*}[h]
    \centering
    \includegraphics[width=\textwidth, height=0.85\textheight, keepaspectratio]{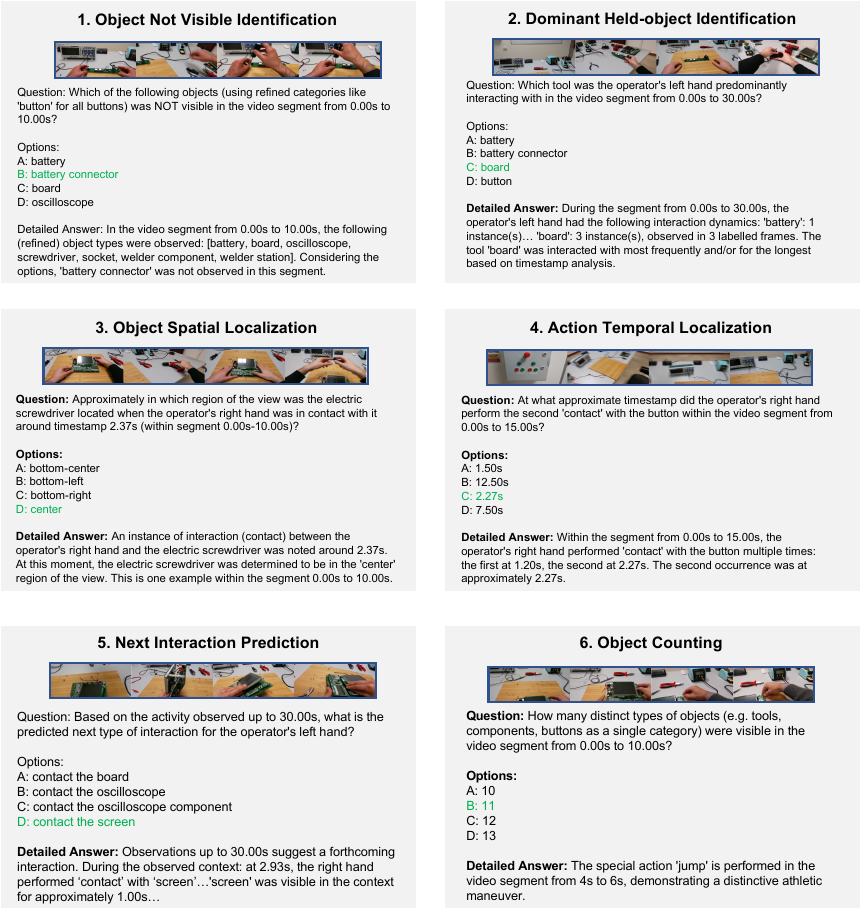} 
    \caption{Representative QA examples from the Industry domain.}
    \label{fig:appendix_industry}
\end{figure*}

\begin{figure*}[h]
    \centering
    \includegraphics[width=\textwidth, height=0.85\textheight, keepaspectratio]{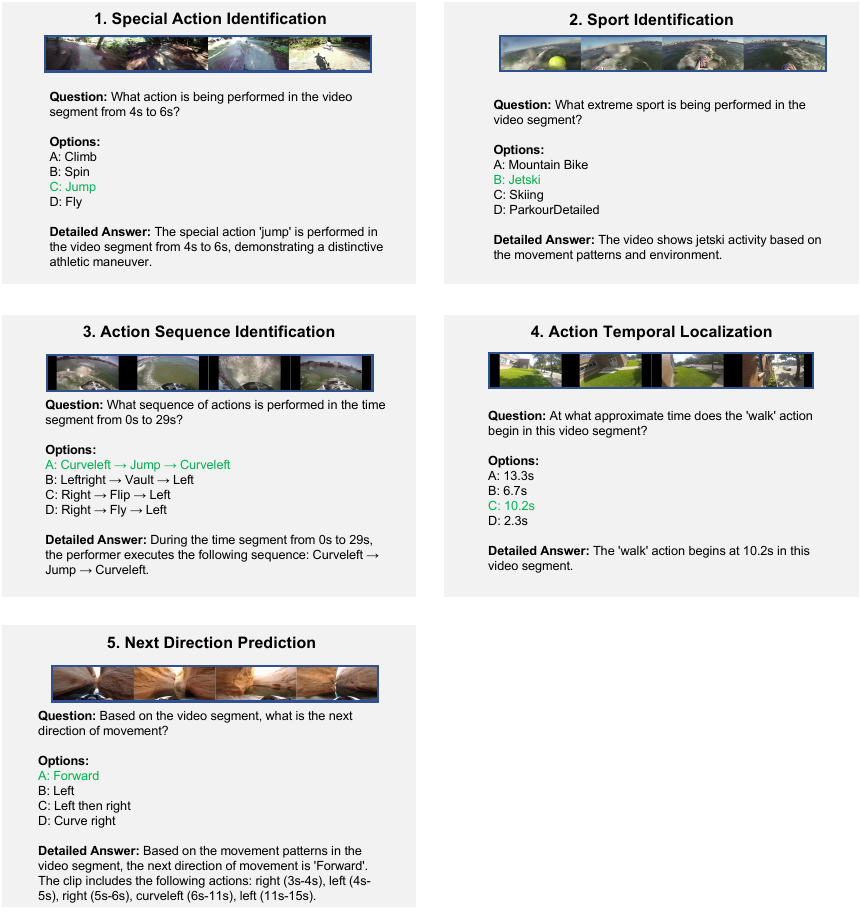} 
    \caption{Representative QA examples from the Extreme Sports domain.}
    \label{fig:appendix_Xsport}
\end{figure*}

\begin{figure*}[h]
    \centering
    \includegraphics[width=\textwidth, height=0.85\textheight, keepaspectratio]{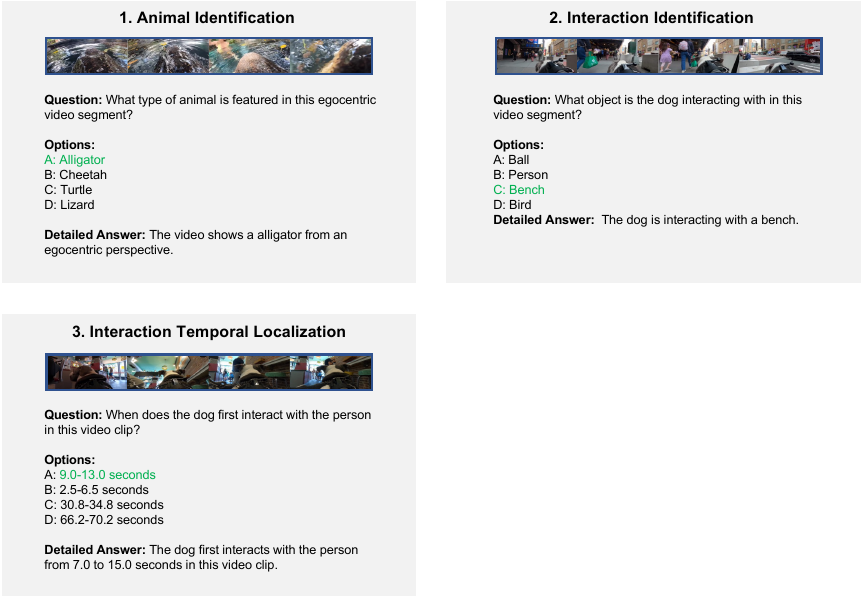} 
    \caption{Representative QA examples from the Animal Perspective domain.}
    \label{fig:appendix_Animal}
\end{figure*}

In the following, we present representative QA examples for all sub-tasks across four distinct domains — \textbf{Surgery} (Figure~\ref{fig:appendix_surgery}), \textbf{Industry} (Figure~\ref{fig:appendix_industry}), \textbf{Extreme Sports} (Figure~\ref{fig:appendix_Xsport}), and \textbf{Animal Perspective} (Figure~\ref{fig:appendix_Animal}).  

